\documentclass[10pt,twocolumn,letterpaper]{article}

\usepackage{cvpr}      
\usepackage{tikz}
\usetikzlibrary{positioning, arrows.meta}
\usepackage[T1]{fontenc} 
\usepackage{iftex}
\usepackage{amsmath}
\usepackage{algorithm}
\usepackage{algpseudocode}
\usepackage{stmaryrd}
\usepackage{float}
\usepackage{xhfill}
\usepackage{bbm}
\usepackage{svg}
\usepackage{multirow}
\usepackage{booktabs}
\usepackage{colortbl}
\usepackage{makecell}
\usepackage{graphicx}
\PassOptionsToPackage{svgnames,x11names,dvipsnames}{xcolor}
\usepackage[most]{tcolorbox}
\usepackage[percent]{overpic}
\usepackage{lipsum}

\definecolor{cvprblue}{rgb}{0.21,0.49,0.74}
\definecolor{LightGrey}{rgb}{0.9,0.9,0.9}

\pdfmapline{=serpantinelightoblique < Serpentine_LightOblique.ttf <T1-WGL4.enc}
\pdfmapline{=serpantinemediumoblique < Serpentine_MediumOblique.ttf <T1-WGL4.enc}
\pdfmapline{=serpantineboldoblique < Serpentine_BoldOblique.ttf <T1-WGL4.enc}

\newcommand\serpantinelight[1]{{\usefont{T1}{serpantinelightoblique}{m}{n} #1 }}
\newcommand\serpantinemedium[1]{{\usefont{T1}{serpantinemediumoblique}{m}{n} #1 }}

\newcommand\tstrut{\rule{0pt}{2.4ex}}
\newcommand\bstrut{\rule[-1.0ex]{0pt}{0pt}}

\newcommand\blfootnote[1]{%
  \begingroup
  \renewcommand\thefootnote{}\footnote{#1}%
  \addtocounter{footnote}{-1}%
  \endgroup
}

\newcommand{\centerrule}[1]{\raisebox{.5ex}{\rule{#1}{.5pt}}}

\newtcbox{\inlinebox}[1][]{enhanced,
 box align=base,
 nobeforeafter,
 colback=white,
 colframe=black,
 size=small,
 left=0pt,
 right=0pt,
 boxsep=2pt,
 #1}

\usepackage[pagebackref,breaklinks,colorlinks,allcolors=cvprblue]{hyperref}

\def\paperID{866} 
\def\confName{CVPR}
\def\confYear{2026}

\title{\serpantinemedium{TRANSPORTER}: Transferring Visual Semantics from VLM Manifolds}

\author{Alexandros Stergiou\\
University of Twente, NL\\
{\tt \href{https://alexandrosstergiou.github.io/project_pages/TRANSPORTER}{https://alexandrosstergiou.github.io/TRANSPORTER}}
}

\begin{document}
\twocolumn[{%
\renewcommand\twocolumn[1][]{#1}%
\maketitle
\begin{center}
    \centering
    \captionsetup{type=figure}
    \includegraphics[width=\textwidth,clip]{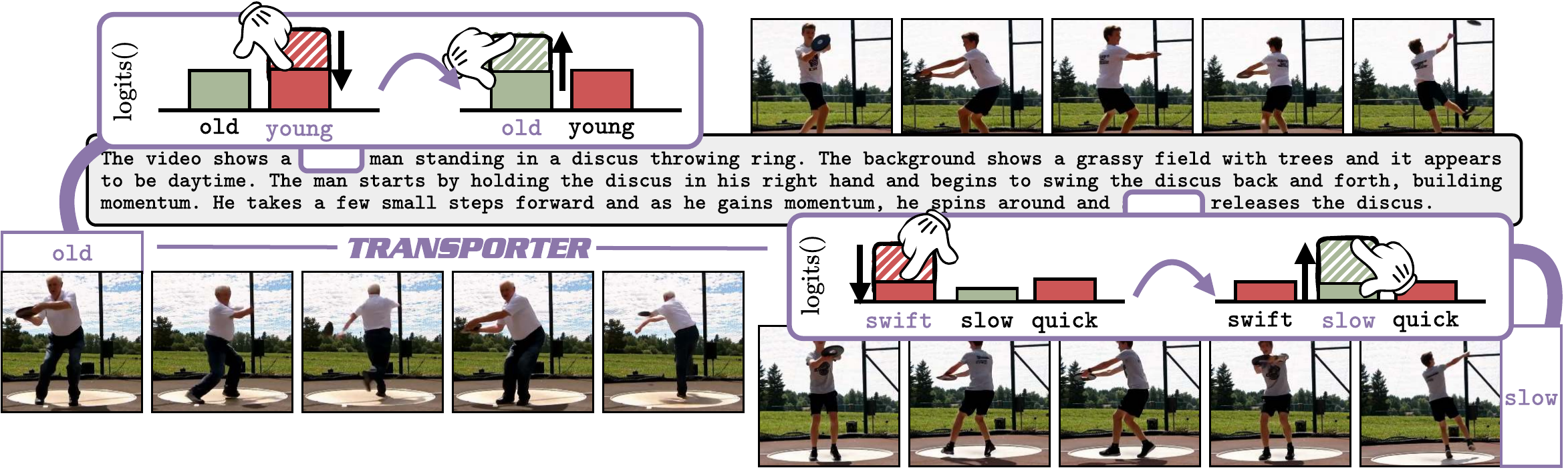}
    \captionof{figure}{\vspace{-.1em}\textbf{Generated videos representing VLM logit modulations with} \serpantinemedium{TRANSPORTER}. Videos that capture different logit predictions are obtained by coupling VLM embeddings to generative representations. Given the generated VLM caption and a target modulation of objects, actions, or scene attributes, \serpantinelight{TRANSPORTER} guides the video generation process to reflect token logit score changes. Embeddings are decoded into logit-score-aligned videos, as shown by the bottom videos that shift \inlinebox{\texttt{young}} for \inlinebox{\texttt{old}} and \inlinebox{\texttt{swift}} for \inlinebox{\texttt{slow}}.}
\label{fig:teaser}

\end{center}%
}]

\maketitle
\begin{abstract}
How do video understanding models acquire their answers? Although current Vision Language Models (VLMs) reason over complex scenes with diverse objects, action performances, and scene dynamics, understanding and controlling their internal processes remains an open challenge. Motivated by recent advancements in text-to-video (T2V) generative models, this paper introduces a logits-to-video (L2V) task alongside a model-independent approach, \serpantinelight{TRANSPORTER}, to generate videos that capture the underlying rules behind VLMs' predictions. Given the high-visual-fidelity produced by T2V models, \serpantinelight{TRANSPORTER} learns an optimal transport coupling to VLM's high-semantic embedding spaces. In turn, logit scores define embedding directions for conditional video generation. \serpantinelight{TRANSPORTER} generates videos that reflect caption changes over diverse object attributes, action adverbs, and scene context. Quantitative and qualitative evaluations across VLMs demonstrate that L2V can provide a fidelity-rich, novel direction for model interpretability that has not been previously explored. 
\end{abstract}
\section{Introduction}
\label{sec:intro}

The world is full of rich visual signals. For example, a discus throw, as in~\cref{fig:teaser}, can include variations in the scene dynamics, people or object attributes, and action performances. Video understanding has experienced drastic growth with detecting, predicting, captioning, grounding, and reasoning actions from encodings~\cite{stergiou2025time}. This convergence from visual attributes to context-rich semantics has led to Vision-Language Models (VLMs)~\cite{team2025gemma,zhang2025videollama,comanici2025gemini,abouelenin2025phi} that address tasks in tandem. Despite significant progress, uncovering explanations behind model decisions has been a longstanding challenge. Existing approaches prompt~\cite{wei2022chain}, linearly probe~\cite{hernandez2024inspecting}, or decode hidden embeddings~\cite{pal2023future,chen2024selfie} to gain text-based descriptions that are often sensitive to changes~\cite{arditi2024refusal}, of limited length, or misrepresent internal processes~\cite{turpin2023language}. To address this gap, this paper revisits visual explanations and presents a logits-to-video (L2V) generative task for synthesizing videos corresponding to logit distributions. As VLMs provide answers as probability distributions over a token dictionary, transporting logits to videos can directly and causally represent relevant insights. 

To model L2V, a VLM-independent approach is introduced. \serpantinelight{TRANSPORTER} creates probabilistic paths between high-semantic and visually-fidelitous representations, allowing for fine-grained generation controlled by VLM predictions (logits). Unlike existing methods that rely on text-based explanations or visual-saliency attributions, \serpantinelight{TRANSPORTER} visually represents modulations across VLM tokens. During training, video attribute pair modulations are encoded to latent vectors. In inference, their logit divergence is used as a condition for video generation. \serpantinelight{TRANSPORTER} provides a novel interactive explainability paradigm that not only verifies the alignment between learned semantics and relevant visual representations, but also explores learned object-attribute correlations.

This work's contributions are: (i) L2V, a novel controlled generation task for VLM visual explanations. (ii) \serpantinelight{TRANSPORTER}, a model that learns optimal transport paths between local geometric relationships and global representations across embedding spaces. (iii) Learnable latents to modulate videos by the target logit divergence. (iv) Empirical evaluations and ablations across settings.
\section{Related works}
\label{sec:related_work}

Approaches for interpreting deep models can be divided into three groups, detailed below.

\noindent
\textbf{Attribution visualizations} relate input contributions to model predictions. For image-based models, approaches have focused on the regional saliency of categories~\cite{chattopadhay2018grad,fong2017interpretable,petsiuk2018rise}, object parts~\cite{bau2017network}, or gradients~\cite{shrikumar2017learning,sundararajan2017axiomatic}. Extensions also propagated local relevance through attention~\cite{bousselham2025legrad}, or layer-averaged activations~\cite{chefer2021generic}, while other efforts visualized attributions in video classification~\cite{stergiou2019saliency,li2021towards}. As recent models represent vision and language in shared embedding spaces~\cite{radford2021learning,zhai2023sigmoid}, approaches have explored text-based attributions with point-wise vision-language mutual information maximization~\cite{hernandez2021natural,bykov2023labeling}, score-based pairwise similarities~\cite{pach2025sparse,gong2025boosting,balasubramanian2024decomposing}, and learned ensemble models~\cite{you2025sum}. Prompt-to-image relevance~\cite{mao2023doubly} has also been visualized through 
image-level object highlights~\cite{shtedritski2023does,moayeri2023text},
\texttt{CLS} token decomposition~\cite{yu2024attention}, and cluster graphs over latent representations~\cite{fel2023holistic,kowal2024visual}. In contrast, \serpantinelight{TRANSPORTER} focuses on visualizing VLM semantics within the challenging video domain by generating attribute-controlled videos.

\noindent
\textbf{Representation decomposition} methods visualize parts of objects and category relations~\cite{fel2023craft}, or decompose activations to vector directions~\cite{graziani2023uncovering}. Vision-language approaches related object semantics to feature~\cite{gandelsman2024interpreting,menon2023visual,jiang2025interpreting}, or learned latent~\cite{chen2023interpreting} descriptions. Works have studied text-to-image correspondence based on embedding distances~\cite{yang2024language,zhang2023diagnosing}, ablated outputs~\cite{neo2025towards}, and modality-specific differences~\cite{hua2025vision,lan2024clearclip,huo2024mmneuron}. Prototype-based approaches~\cite{sui2025just,du2022weakly,wang2023learning,xue2022protopformer} have gained traction with contrastive-learned embeddings~\cite{du2022weakly}, semantics-caching-instance-shifting representations~\cite{sui2025just}, and semantic boundary centering~\cite{wang2023learning,xue2022protopformer}. Optimal Transport (OT) has also been used to align visual features to text semantics~\cite{chen2023plot,zhu2024awt}, and cluster features based on activations~\cite{wang2025pot,zhao2021towards}. This paper explores a different direction. It uses OT to couple high-semantic embeddings with detailed visual representations for video explanations.

\noindent
\textbf{Concept-based} methods aim to invert model inputs~\cite{simonyan2013deep,nguyen2016multifaceted,yin2020dreaming,hatamizadeh2022gradvit}. Most objectives update inputs with Activation Maximization (AM)~\cite{simonyan2013deep} to increase activations of neurons~\cite{nguyen2016synthesizing}, class scores~\cite{nguyen2016multifaceted}, gradients~\cite{hatamizadeh2022gradvit}, features~\cite{olah2020zoom,fel2023unlocking}, or language semantics~\cite{jain2025mimic}. AM has also been adapted for convolution~\cite{feichtenhofer2020deep} and attention~\cite{stergiou2023leaping} video models. Due to parameter sensitivity~\cite{yin2020dreaming,stergiou2023leaping,ghiasi22plug}, more recent works instead trained~\cite{koh2020concept} or fine-tuned~\cite{laguna2024beyond} copies of models for visually interpretable features~\cite{koh2020concept}, semantic-alignment~\cite{srivastava2024vlg,kulkarni2025interpretable}, or joint vision-language representations~\cite{d2025implicit,fort2025direct}. Concept editing has been explored for T2I models~\cite{gandikota2024concept,hertz2022prompt,baumann2025continuous,parihar2024precisecontrol}. Approaches employed low-ranking adapters~\cite{hu2022lora,gandikota2024concept}, prompt-editing~\cite{hertz2022prompt}, and prompt-pair embedding differences~\cite{baumann2025continuous,parihar2024precisecontrol} for control over the image generation. Drawing inspiration from concept editing in T2I, the proposed method uses a control-based objective to generate videos that visually represent VLM tokens (L2V).

\begin{figure*}[t]
\centering
\begin{subfigure}[b]{\textwidth}
    \centering
    \includegraphics[width=\textwidth]{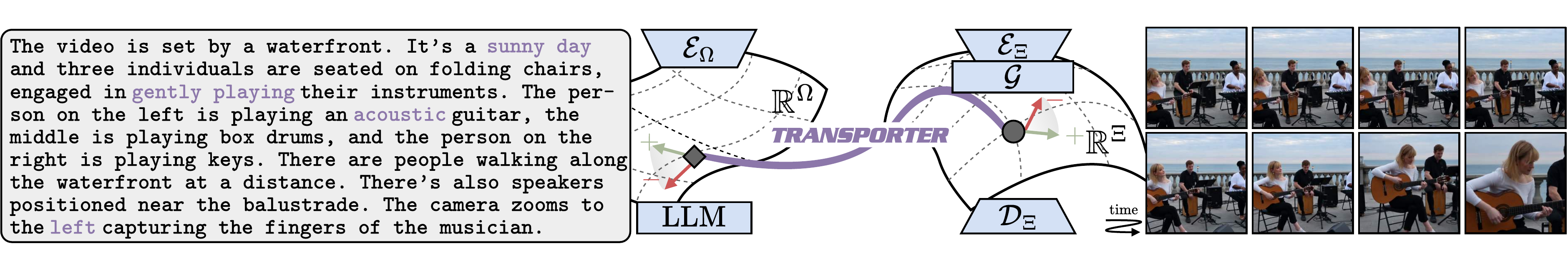}
    \caption{\textbf{L2V overview}}
    \label{fig:transporter_main_a}
\end{subfigure}\\
\begin{subfigure}[b]{0.27\textwidth}
    \centering
    \begin{overpic}[width=\textwidth]{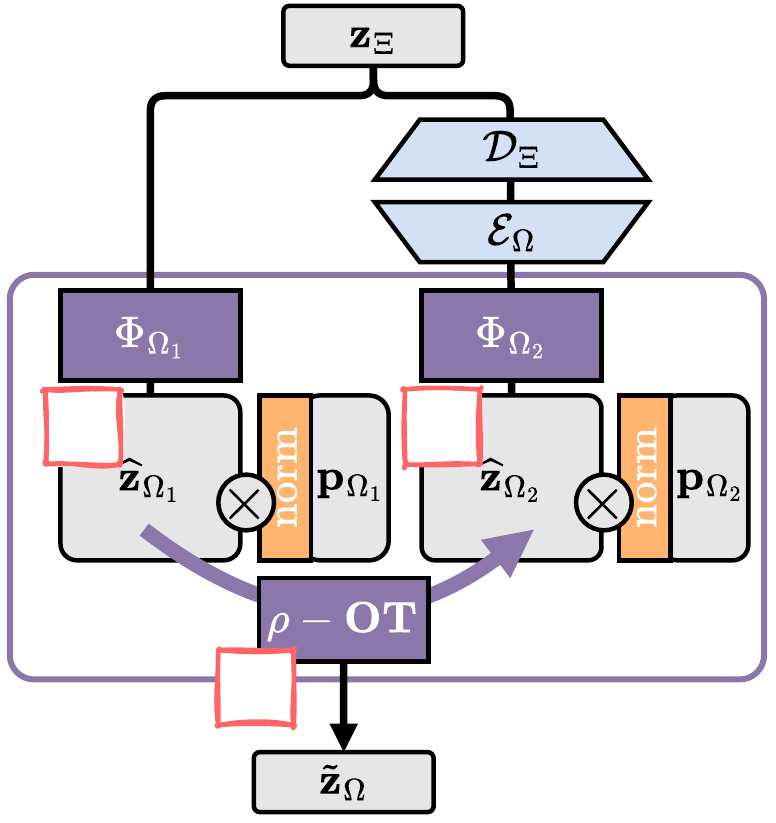}
     \put(6.5,46.4){\small(\ref{eq:mse})}
     \put(50.1,46.6){\small(\ref{eq:gram})}
     \put(27.6,14.8){\small(\ref{eq:pOT})}
     \end{overpic}
    \caption{\textbf{Coupling network training} $\Phi$}
    \label{fig:transporter_main_b}
\end{subfigure}
\tikz{\draw[densely dashed, thick](0,5.5) -- (0,0);}
\begin{subfigure}[b]{0.29\textwidth}
    \centering
    \begin{overpic}[width=\textwidth]{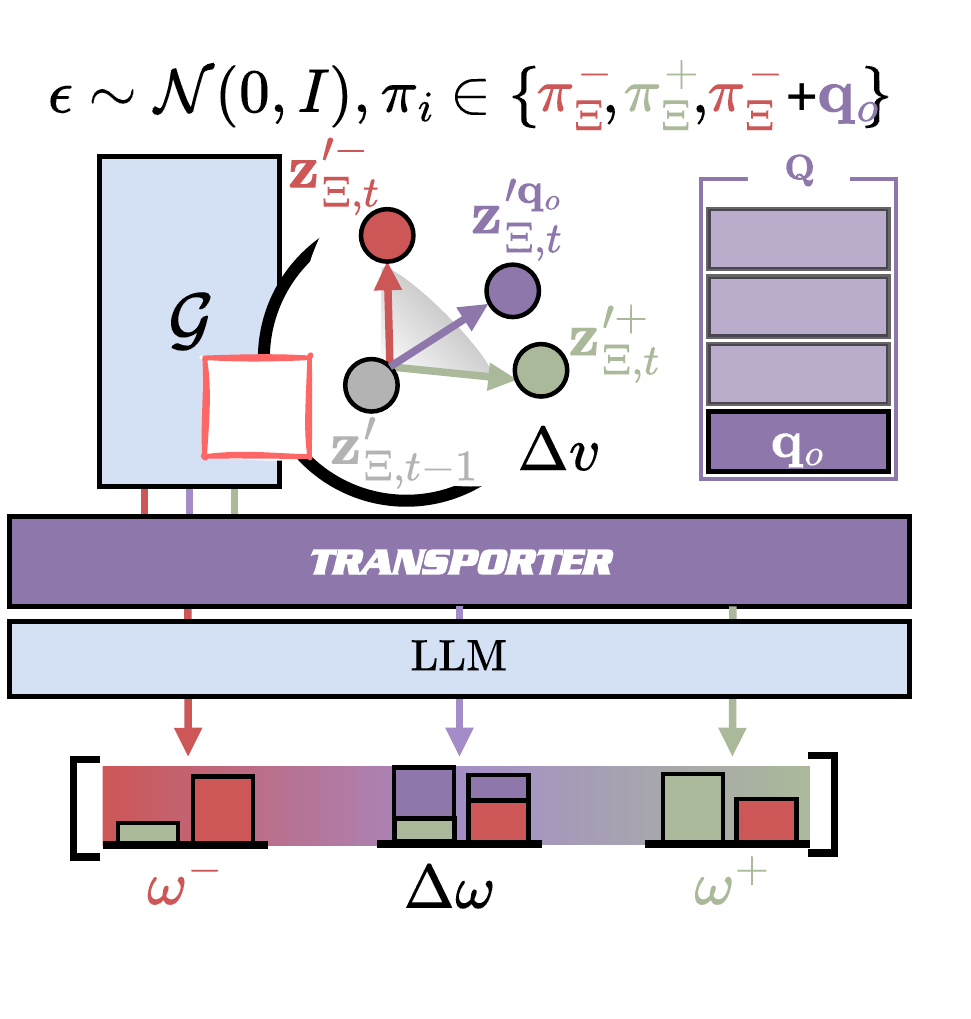}
     \put(21.75,58.2){\small(\ref{eq:bank})}
     \end{overpic}
    \caption{\textbf{Concept bank} $\mathbf{Q}$ \textbf{training}}
    \label{fig:transporter_main_c}
\end{subfigure}
\tikz{\draw[densely dashed, thick](0,5.5) -- (0,0);}
\begin{subfigure}[b]{0.40\textwidth}
    \centering
    \includegraphics[width=\textwidth]{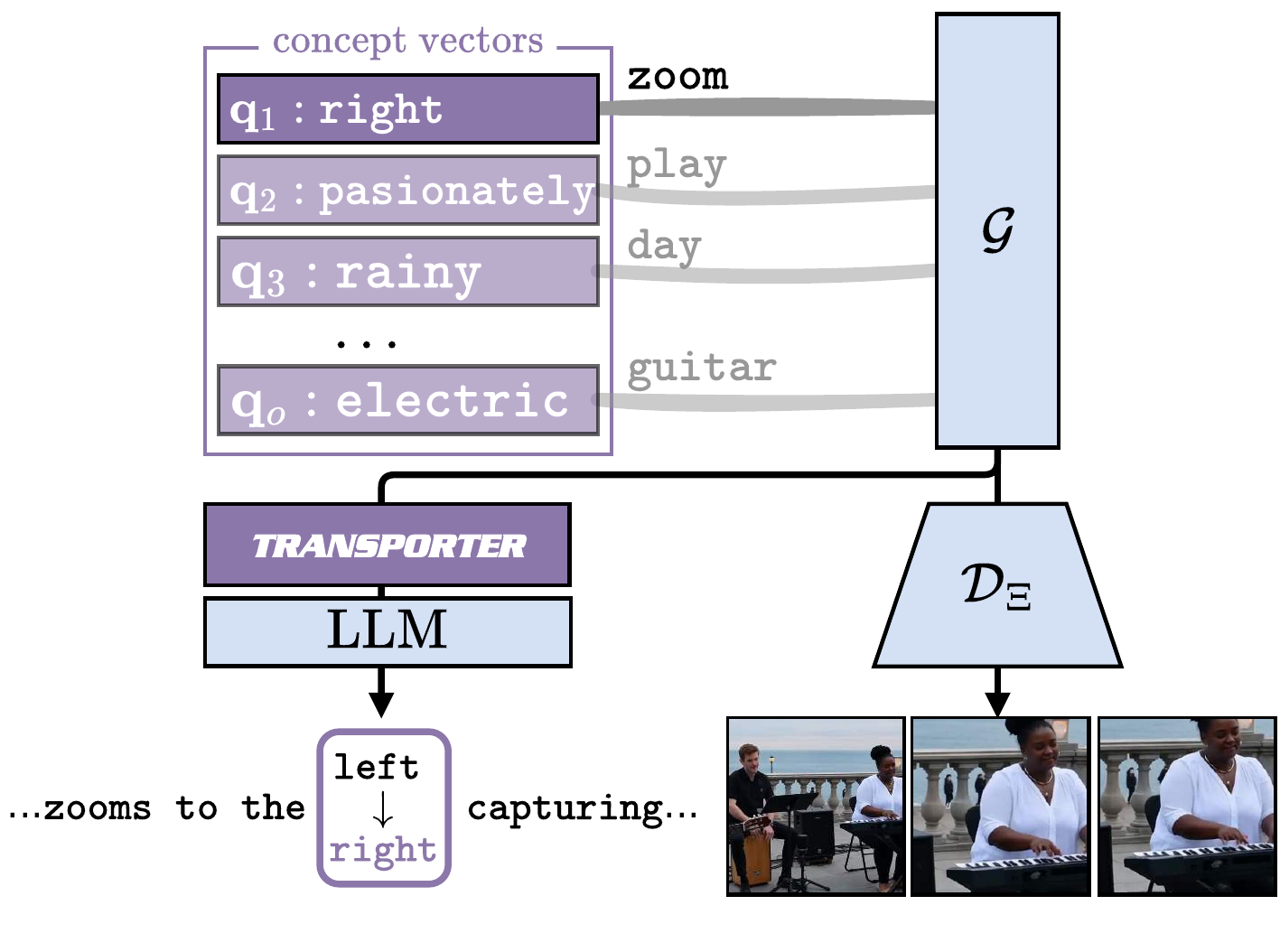}
    \caption{\textbf{Inference step}}
    \label{fig:transporter_main_d}
\end{subfigure}
\caption{(a) \textbf{L2V with} \serpantinemedium{TRANSPORTER}: Embeddings $\mathbf{z}_\Xi \in \mathbb{R}^\Xi$ are \textbf{coupled with network} $\Phi$ and \textbf{concept bank} $\mathbf{Q}$. (b) \textbf{Coupling network} $\Phi$ initially projects $\mathbf{z}_\Xi$ with condition $\pi_\Xi$ to $\widehat{\mathbf{z}}_{\Omega_1}=\Phi_{\Omega_1}(\mathbf{z}_\Xi,\pi_\Xi)$. Latents $\widehat{\mathbf{z}}_{\Omega_2} \in \mathbb{R}^\Omega$ are obtained with $\Phi_{\Omega_2}$ over decoder $\mathcal{D}_\Xi$ and encoder $\mathcal{E}_\Omega$ latents. The Learnable Optimal Transport ($\rho$-OT) module uses updatable projection vectors $\mathbf{p}_{\Omega_1},\mathbf{p}_{\Omega_2}$ to transport embeddings to $\tilde{\mathbf{z}}_\Omega$. The divergence between pairs ${\pi^{-}},{\pi^{+}}$ is used to train the (c) \textbf{Concept bank} $\mathbf{Q}=\{\mathbf{q}_o:o\in \mathcal{O}\}$ given the probability path difference $\Delta v$ between conditions,  weighted by the LLM logit distributions change $\Delta \omega$ for $\omega^{-}$ to $\omega^{+}$. For each (d) \textbf{Inference step}, latents $\mathbf{q}_o$ are added to conditions to transport noise latent $\mathbf{\epsilon}\sim \mathcal{N}(0,\mathbf{I})$ and generate videos and captions.}
\label{fig:transporter_main}
\vspace{-1em}
\end{figure*}

\section{Method}
\label{sec:method}

This section provides preliminary definitions for T2V and formulates the L2V objective for explanations~(\cref{sec:method::definition}); introduces embedding manifold coupling~(\cref{sec:method::description}); and the learnable concept bank~(\cref{{sec:method::bank}}), which together form \serpantinelight{TRANSPORTER}; and concludes with an overview of a inference pass over all modules~(\cref{sec:method::logits}).

\subsection{Definitions}
\label{sec:method::definition}

\noindent
\textbf{Preliminaries}. Recent T2V models~\cite{wan2025wan,jin2025pyramidal} are trained with Conditional Flow Matching (CFM)~\cite{lipman2023flow,albergo2023building,liu2023flow,tong2024improving} to model velocity fields between standard Gaussian priors $\epsilon\! \sim\!\! \mathcal{N}(0,\mathbf{I})$ and target distributions of (latent, condition) pairs $(\mathbf{z}_\Xi,\pi_\Xi)$, where $\mathbf{z}_\Xi\! \in\! \mathbb{R}^\Xi$ are $\mathbf{N}$-token embeddings of video $\mathbf{x}$ and $\pi_\Xi = \mathcal{T}_\Xi(\pi)$ is a tokenized text condition $\pi$. Probability paths $\mathbf{z}'_{\Xi,t}$ for time $t\in[0,1]$ are commonly constructed with a linear interpolation $\mathbf{z}'_{\Xi,t}=t\mathbf{z}_\Xi+(1-t)\epsilon$ and conditional velocity field $\mathit{v}(\mathbf{z}'_{\Xi,t}|\pi_\Xi) \triangleq  \mathbf{z}_\Xi - \epsilon$. Network $\mathcal{G}$ is trained to regress the conditional velocity across sampled paths with Mean Squared Error (MSE):
\begin{equation}
    \mathcal{L}_{CFM} = \mathbb{E}_{\epsilon,t,(\mathbf{z}_\Xi,\pi_\Xi)}\|\mathit{v}(\mathbf{z}'_{\Xi,t}|\pi_\Xi)-\mathcal{G}(\mathbf{z}'_{\Xi,t},\pi_\Xi,t)\|^2
\label{eq:cfm}
\end{equation}
\noindent

\noindent
\textbf{L2V}, shown in~\cref{fig:transporter_main_a}, considers VLM predicted logits $\omega\!=\!\texttt{logit}(\pi_\Omega)$ for caption $\pi_\Omega\!=\!\text{LLM}(\beta,\mathbf{z}_\Omega)$ from text query $\beta$, and video encodings $\mathbf{z}_\Omega\!=\!\mathcal{E}_\Omega(\mathbf{x}) \in \mathbb{R}^\Omega$ with $\mathbf{N}$ tokens. Captions include an arbitrary number of subjects and attributes. L2V aims to generate video $\mathbf{x}'$ with $\Delta\omega$ for the change between token logits $\omega^{-}$ from $\pi_\Omega^-$ to $\omega^{+}$ from $\pi_\Omega^+$. For example, given $\omega^{-}$ logit for \inlinebox{\texttt{happy}}, the goal is to generate video $\mathbf{x}'$ to instead match $\omega^{+}$ logit for \inlinebox{\texttt{sad}}.

\subsection{Coupling visual fidelity to semantics}
\label{sec:method::description}

Core to the paper's proposal is the coupling of latent representations between the generator's $\mathbb{R}^\Xi$ and VLM's $\mathbb{R}^\Omega$ spaces. As embeddings $\mathbf{z}_\Xi$ can be decoded back to video by the generator's (variational) decoder $\mathcal{D}_\Xi$, a natural choice for L2V would be to decode-and-re-encode $\mathbf{z}_{\Xi\rightarrow\Omega}=\mathcal{E}_{\Omega}(\mathcal{D}_\Xi(\mathbf{z}_\Xi))$. Although the reconstructed embeddings are within $\mathbb{R}^\Omega$'s manifold, the VAE decoder introduces stochastic variation~\cite{kingma2014auto}. The resulting token dynamics differ from deterministically encoded encodings $\mathbf{z}_\Xi,\mathbf{z}_\Omega$. Thus, as overviewed in~\cref{fig:transporter_main_b}, a coupling network $\Phi$ is used to project and transfer $\mathbf{z}_\Xi$ to  $\tilde{\mathbf{z}}_\Omega\!\approx\!\mathbf{z}_\Omega$ approximating the representations within $\mathbb{R}^\Omega$.

\begin{figure*}[t]
    \centering
    \includegraphics[width=\linewidth]{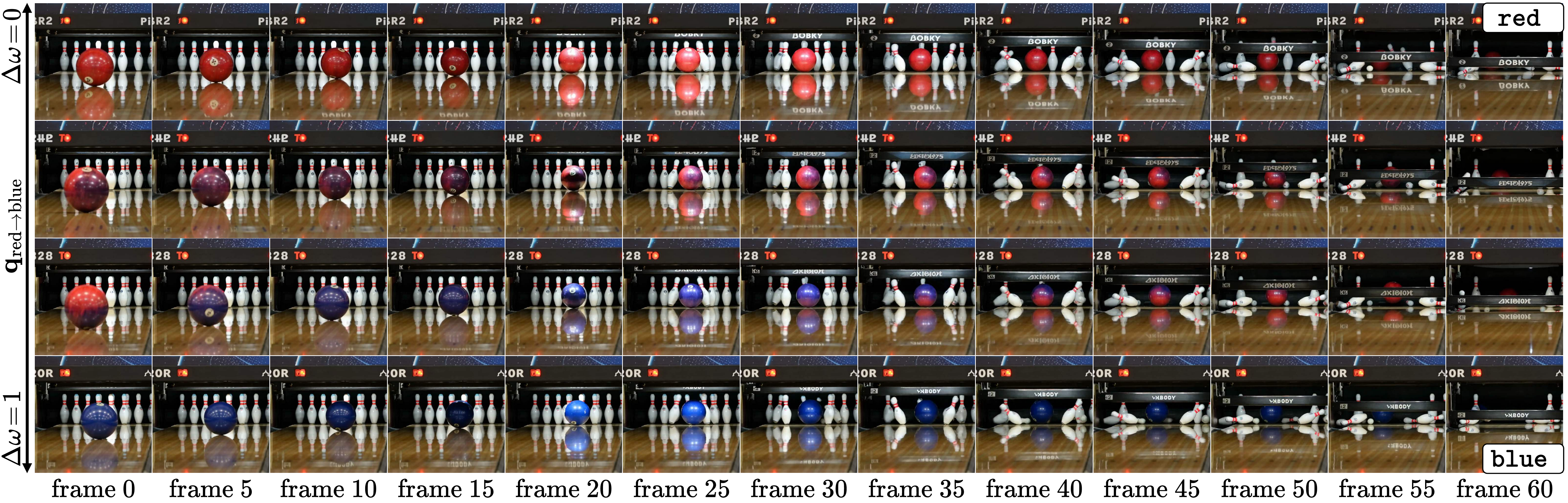}
    \caption{\textbf{Concept attribute control with} \serpantinemedium{TRANSPORTER} given the caption: \texttt{A close up shot of a} \inlinebox{\textcolor{white}{attr}} \texttt{bowling ball hitting the pins in a bowling alley}. Initially, \inlinebox{\texttt{red}} is used to obtain generator/VLM encodings $\pi^-_\Xi,\pi^-_\Omega$. Vector $\mathbf{q}_{\text{red}\rightarrow\text{blue}}$ is added to $\pi^-_\Xi$ based on divergence $\Delta \omega$ to $\pi^+\!\!=$\inlinebox{\texttt{blue}}. As shown, the generated videos are of high visual fidelity while also preserving scene dynamics across modulations; \textit{e.g.}, camera view, object motions and their affordances over time, as well as time-relevant interactions. Changes are only seen specifically for the target attribute. Vector $\mathbf{q}_{\text{red}\rightarrow\text{blue}}$ used is learned with $\Delta \omega$ from Gemma 3~\cite{team2025gemma} logits.}
    \label{fig:deltaomega_bowling}
    \vspace{-1em}
\end{figure*}

\noindent
\textbf{Projecting to target space}. Embeddings $\mathbf{z}_\Xi$ are first projected to $\widehat{\mathbf{z}}_{\Omega_1} \!\! = \! \Phi_{\Omega_1}(\mathbf{z}_\Xi,\pi_\Xi) \! \in \! \mathbb{R}^\Omega$ by $\Phi_{\Omega_1}$. The module optimizes a MSE objective~(\ref{eq:mse}) between obtained projections $\widehat{\mathbf{z}}_{\Omega_1}$ and target encodings $\mathbf{z}_\Omega$.

\noindent
\textbf{Attending token structure}. As the local geometric relationships between tokens are not directly addressed by (\ref{eq:mse}), $\mathbf{z}_{\Xi \rightarrow \Omega}$ are used to learn soft targets $\widehat{\mathbf{z}}_{\Omega_2}\!=\!\Phi_{\Omega_2}(\mathbf{z}_{\Xi \rightarrow \Omega})\! \in \! \mathbb{R}^\Omega$. Module $\Phi_{\Omega_2}$ optimizes a Gram-matrix loss~\cite{gatys2016image,li2017demystifying} to match $\mathbf{z}_\Omega$'s relational token structure~(\ref{eq:gram}).

\noindent
\begin{minipage}{0.44\linewidth}
\begin{equation}
    \mathcal{L}_{\Phi_{\Omega_1}}\!\!=\!\! \| \mathbf{z}_\Omega \! - \! \widehat{\mathbf{z}}_{\Omega_1} \|^2
\label{eq:mse}
\end{equation}
\end{minipage}%
\hfill
\begin{minipage}{0.55\linewidth}
\begin{equation}
    \mathcal{L}_{\Phi_{\Omega_2}} \!\! = \!\! \|\mathcal{H}(\mathbf{z}_\Omega) \! - \! \mathcal{H}(\widehat{\mathbf{z}}_{\Omega_2}) \|^2
\label{eq:gram}
\end{equation}
\end{minipage}
\noindent
where $\mathcal{H}(\mathbf{z}_\Omega), \mathcal{H}(\widehat{\mathbf{z}}_{\Omega_2})$ are the $\mathbb{R}^{|\mathbf{N}| \times |\mathbf{N}|}$ Gram matrices computed as the inner product between tokens from $\mathbf{z}_\Omega,\widehat{\mathbf{z}}_{\Omega_2}$ respectively to match token structures within embeddings.

\noindent
\textbf{Learnable OT}. Given globally projected $\widehat{\mathbf{z}}_{\Omega_1}$ and local structure-preserving $\widehat{\mathbf{z}}_{\Omega_2}$, their complementary representation properties are combined to distilled embeddings $\tilde{\mathbf{z}}_\Omega$. Following recent works on distribution matching over varying-size embedding spaces~\cite{titouan2019sliced,piening2025novel}, a novel learnable entropic-based OT ($\rho$-OT) module is defined. Given $\widehat{\mathbf{z}}_{\Omega_1}$ and $\widehat{\mathbf{z}}_{\Omega_2}$, $\rho$-OT uses $\{\mathbf{p}_{\Omega_1,\rho}\}_{\rho=1}^P$ and $\{\mathbf{p}_{\Omega_2,\rho}\}_{\rho=1}^P$ sets of $P$ learnable projection vectors. Each $\mathbf{p}_{\cdot,\rho} \! \in \! \mathbb{R}^{\Omega}$  projects tokens $i,j \! \in \!\! \mathbf{N}$ onto scalars for each projection $\rho \! \in \!\! \{1,\dots,P\}$:
\begin{align}
    \mathbf{a}_{i,\rho} &= \langle \widehat{\mathbf{z}}_{\Omega_1,i},\mathbf{p}_{\Omega_1,\rho} \rangle \forall i \in \mathbf{N}\\
    \mathbf{b}_{j,\rho} &= \langle \widehat{\mathbf{z}}_{\Omega_2,j},\mathbf{p}_{\Omega_2,\rho} \rangle \forall j \in \mathbf{N}
\label{eq:proj}
\end{align}
For each $\rho$, transport plan $\tilde{\mathbf{T}}$ is found by minimizing transport cost $\mathbf{M}_{i,j,\rho}\! =\! \|\mathbf{a}_{i,\rho}\!-\!\mathbf{b}_{j,\rho}\|_2$, with entropic regularization controlled by temperature $\tau$. The full continuous and discrete formulations are expanded in \S\ref{sec:entropic_ot}. As solving the \emph{full} doubly-constrained problem is computationally intensive, an efficient, closed-form approximation with \emph{partial} double-stochastic iterations is used to regress $P$-averaged optimal transport plans from $\mathbf{T}_{i,j,\rho} \! \propto \! \text{exp}(-\mathbf{M}_{i,j,\rho}/\tau)$. The resulting $\tilde{\mathbf{T}}\! \in \! \mathbb{R}^{|\mathbf{N}|\times|\mathbf{N}|}$ can be applied to obtain $\tilde{\mathbf{z}}_\Omega \!= \!\tilde{\mathbf{T}}\widehat{\mathbf{z}}_{\Omega_1}$. The projection vectors $\mathbf{p}_{\Omega_1},\mathbf{p}_{\Omega_2}$ are updated based on a joint (\ref{eq:mse}) and (\ref{eq:gram}) objective:
\begin{equation}
    \mathcal{L}_{\rho\text{-OT}} = \| \mathbf{z}_\Omega - \tilde{\mathbf{z}}_\Omega \|^2 + \|\mathcal{H}(\mathbf{z}_\Omega) - \mathcal{H}(\tilde{\mathbf{z}}_\Omega) \|^2
\label{eq:pOT}
\end{equation}
A detailed algorithmic formulation of the process is available in \S \ref{sec:rho_ot_formulation} of the supplementary material.

\begin{table*}[t]
\begin{minipage}[b]{0.49\textwidth}%
\caption{\textbf{Quantitative comparisons on VideoLLaMA 3, Gemma 3, Phi 4 MM}. FVD and $\text{LPIPS}^{\textcolor{red}{\text{v}}}$ evaluate the visual quality of generated videos in comparison to videos from VidChapters7M~\cite{yang2023vidchapters}. $\text{CLIP}^{\textcolor{red}{\text{v}}}$ score, aes, and $\Delta$ evaluate condition alignment. AM-optimized approaches are adjusted to both video data and VLMs by maximizing either $\omega^{\!+}$ or $\omega^{\!-}\!\! +\!\! \Delta\omega$. Top performances are in \textbf{bold}.}
\label{tab:method_comparisons}
\end{minipage}
\hfill
\begin{subtable}[t]{0.49\textwidth}
    \vspace*{-7.5em}
    \setcounter{subtable}{1}
    \caption{Gemma 3}
    \centering
    \resizebox{\linewidth}{!}{%
    \setlength\tabcolsep{3.5pt}
    \begin{tabular}{l c cc c c}
        \toprule
         \multirow{2}{*}{Method} & \multirow{2}{*}{$\text{FVD}\!\downarrow$} &  \multirow{2}{*}{$\text{LPIPS}^{\textcolor{red}{\text{v}}}\!\downarrow$} & \multicolumn{3}{c}{$\text{CLIP}^{\textcolor{red}{\text{v}}}$} \\ \cline{4-6}
         & & & score$\uparrow$ & aes.$\uparrow$ & $\Delta\!\uparrow$ \tstrut \\
         \multicolumn{6}{l}{\centerrule{.5cm} Feature vis. maximizing $\omega^{+}$ \centerrule{3.8cm}}\bstrut\\
         Baseline~\cite{stergiou2023leaping} & 2.18e$^3$ & 4.55 & 17.41 & 2.67 & 1.82 \\
         \multicolumn{6}{l}{\centerrule{.5cm} Feature vis. maximizing varying $\omega^{+}+\Delta \omega$ \centerrule{1.6cm}}\bstrut\\
         Baseline~\cite{stergiou2023leaping} & 2.42e$^3$ & 4.72 & 14.13 & 2.05 & 1.89 \\
         \midrule
         \serpantinelight{TRANSPORTER}$\;$ & \textbf{1.05e}$^{\mathbf{2}}$ & \textbf{1.43} & \textbf{36.18} & \textbf{4.21} & \textbf{11.56} \\
         \bottomrule
    \end{tabular}
    }
    \label{tab:method_comparisons::gemma3}
\end{subtable}\\
\begin{subtable}[b]{0.49\textwidth}
    \vspace*{-3.5em}
    \setcounter{subtable}{0}
    \caption{VideoLLaMA 3}
    \centering
    \resizebox{\linewidth}{!}{%
    \setlength\tabcolsep{1.2pt}
    \begin{tabular}{l c cc ccc}
        \toprule
         \multirow{2}{*}{Method} & 
         \multirow{2}{*}{Opt} &
         \multirow{2}{*}{$\text{FVD}\!\downarrow$} & 
         \multirow{2}{*}{$\text{LPIPS}^{\textcolor{red}{\text{v}}}\!\downarrow$} &
         \multicolumn{3}{c}{$\text{CLIP}^{\textcolor{red}{\text{v}}}$} \\ \cline{5-7}
         & & & & score$\uparrow$ & aes.$\uparrow$ & $\Delta\!\uparrow$ \tstrut \\
         \multicolumn{7}{l}{\centerrule{.5cm} Feature vis. maximizing $\omega^{+}$ \centerrule{3.6cm}}\bstrut\\
         AM\textcolor{red}{$^{\text{v}}$}~\cite{simonyan2013deep} & \multirow{4}{*}{AM} & 5.18e$^{3}$ & 6.83 & 9.48 & 1.36 & 1.06 \\
         GradViT\textcolor{red}{$^{\text{v}}$}~\cite{hatamizadeh2022gradvit} & & 2.61e$^3$ & 5.46 & 10.11 & 1.55 & 1.73 \\
         MACO\textcolor{red}{$^{\text{v}}$}~\cite{fel2023unlocking} & & 2.50e$^3$ & 4.52 & 14.23 & 2.27 & 1.56 \\
         LEAPS~\cite{stergiou2023leaping} & & 1.85e$^3$ & 4.37 & 16.74 & 2.56 & 2.28 \\
         \multicolumn{7}{l}{\centerrule{0.5cm} Feature vis. maximizing varying $\omega^{+}+\Delta \omega$ \centerrule{1.4cm}}\bstrut\\
         MACO\textcolor{red}{$^{\text{v}}$}~\cite{fel2023unlocking} & \multirow{2}{*}{AM} & 3.88e$^3$ & 5.85 & 11.54 & 1.91 & 1.78 \\
         LEAPS~\cite{stergiou2023leaping} & & 2.39e$^3$ & 5.12 & 11.95 & 2.13 & 2.30 \\
         \midrule
         \serpantinelight{TRANSPORTER}$\;$ & L2V & \textbf{1.25e}$^{\mathbf{2}}$ & \textbf{1.67} & \textbf{35.44} & \textbf{4.28} & \textbf{12.62} \\
         \bottomrule
    \end{tabular}
    }
    \label{tab:method_comparisons::videollama3}
\end{subtable}
\hfill
\begin{subtable}[b]{0.49\textwidth}
    \vspace{-1em}
    \setcounter{subtable}{2}
    \caption{Phi 4 MM}
    \centering
    \resizebox{\linewidth}{!}{%
    \setlength\tabcolsep{3.5pt}
    \begin{tabular}{l c cc c c}
        \toprule
         \multirow{2}{*}{Method} & \multirow{2}{*}{$\text{FVD}\!\downarrow$} &  \multirow{2}{*}{$\text{LPIPS}^{\textcolor{red}{\text{v}}}\!\downarrow$} & \multicolumn{3}{c}{$\text{CLIP}^{\textcolor{red}{\text{v}}}\!\uparrow$} \\ \cline{4-6}
         & & & score$\uparrow$ & aes.$\uparrow$ & $\Delta\!\uparrow$ \tstrut \\
         \multicolumn{6}{l}{\centerrule{.5cm} Feature vis. maximizing $\omega^{+}$ \centerrule{3.8cm}}\bstrut\\
         Baseline~\cite{stergiou2023leaping} & 2.34e$^3$ & 5.12 & 15.06 & 2.24 & 1.73 \\
         \multicolumn{6}{l}{\centerrule{.5cm} Feature vis. maximizing varying $\omega^{+}+\Delta \omega$ \centerrule{1.6cm}}\bstrut\\
         Baseline~\cite{stergiou2023leaping} & 2.58e$^3$ & 5.43 & 10.45 & 1.87 & 1.54 \\
         \midrule
         \serpantinelight{TRANSPORTER}$\;$ & \textbf{1.42e}$^{\mathbf{2}}$ & \textbf{1.54} & \textbf{35.71} & \textbf{4.18} & \textbf{11.35} \\
         \bottomrule
    \end{tabular}
    }
    \label{tab:method_comparisons::phi4}
\end{subtable}
\vspace{-.9em}
\end{table*}
\setcounter{table}{1}

\subsection{Concept bank learning}
\label{sec:method::bank}

The goal of L2V is to generate embeddings $\mathbf{z}'_\Xi$ that capture VLM logit score changes. This requires identifying latent directions. Given pairs of semantically contrastive VLM tokens, \textit{e.g.}, $\pi^-\!\!\!=$\inlinebox{\texttt{baseball hit}}, $\pi^+\!\!\!=$\inlinebox{\texttt{baseball miss}}, a learnable concept bank $\mathbf{Q}=\{\mathbf{q}_o : o \in \mathcal{O}\}$ consisting of $|\mathcal{O}|$ concepts is created (\cref{fig:transporter_main_c}). Each concept vector $\mathbf{q}_o \in \mathbb{R}^\Xi$, encodes the latent direction of pair $\pi^{-},\pi^{+}$, in turn tokenized to $\pi^-_\Xi,\pi^+_\Xi \in \mathbb{R}^\Xi$. In addition to video embeddings, the condition pair is passed to coupling network $\Phi$ (\cref{fig:transporter_main_b}) so attribute modulations can be represented across both $\mathbb{R}^\Xi$ and $\mathbb{R}^\Omega$ spaces.

\noindent
\textbf{Generator modulations}. Initially, a video caption containing concept $\pi^-$ is tokenized to $\pi^-_\Xi = \mathcal{T}_\Xi(\pi^-)$. Starting from a Gaussian noise latent prior $\epsilon \sim \mathcal{N}(0,\mathbf{I})$, generator $\mathcal{G}$ is used to approximate denoised embedding $\mathbf{z}'^-_{\Xi}$ given condition $\pi^-_\Xi$ over $\bar{t}$ steps. At step $t-1$, the probability path is approximated as:
\begin{equation}
\mathbf{z}'_{\Xi,t} = \mathbf{z}'_{\Xi,t-1}+ (1/\bar{t})\mathcal{G}(\mathbf{z}'_{\Xi,t-1},\pi_\Xi^-,t-1)
\label{eq:gen_pre}
\end{equation}
\noindent
where $\mathbf{z}'_{\Xi,0}\!=\! \epsilon$ at timestep $t\!=\!0$ and $\mathbf{z}'_{\Xi,\bar{t}} \approx \mathbf{z}'^-_{\Xi}$ at $t\!=\!\bar{t}$.

\begin{figure}[t]
    \centering
    \includegraphics[width=\linewidth]{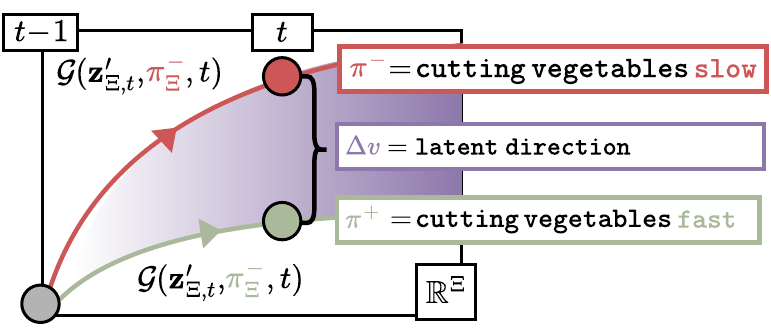}
    \caption{\textbf{Flow path modulation}. Given latents $\mathbf{z}'_{\Xi,t}$ two velocity fields are predicted for conditions $\pi^-,\pi^+$ at step $t$. Their latent divergence $\Delta v$ corresponds to concept/attribute directions.}
    \label{fig:delta_velocity}
    \vspace{-1em}
\end{figure}

At next step $t$, condition $\pi^-_{\Xi}$ can be instead adjusted to include one or multiple subject(s)-specific attribute changes $\pi^+_{\Xi}$. The different velocity fields can be predicted for the adjusted conditions as shown in~\cref{fig:delta_velocity}. The divergence between the two predicted probability paths, $\Delta v = \mathcal{G}(\mathbf{z}'_{\Xi,t},\pi_\Xi^+,t) - \mathcal{G}(\mathbf{z}'_{\Xi,t},\pi_\Xi^-,t)$ corresponds to the difference across the target pairs of concept/attribute modulations. Comparably to finding latent directions in diffusion models~\cite{baumann2025continuous}, the CFM loss~(\ref{eq:cfm}) can be adjusted to include concept vector $\pi_\Xi^-+\delta\mathbf{q}_o$:
\begin{align}
\begin{split}
    \mathcal{L}_{\mathbf{q}_o} = \mathbb{E}_{\epsilon,t,(\mathbf{z}'^{-}_{\Xi},\pi_\Xi^-)}\| &\texttt{sg}\bigl( \mathcal{G}(\mathbf{z}'_{\Xi,t},\pi_\Xi^-,t)+ \delta\Delta v \bigr) \\
    &- \mathcal{G}(\mathbf{z}'_{\Xi,t},\pi_\Xi^-+\delta\mathbf{q}_o,t) \|^2
\label{eq:bank}
\end{split}
\end{align}

\noindent
with $\delta$ control parameter and $\texttt{sg}$ stop-gradient. For both $\pi_\Xi^-$ and $\pi_\Xi^-+\delta\mathbf{q}_o$, generator $\mathcal{G}$ weights remain frozen.

\noindent
\textbf{VLM modulations}. Path divergence $\Delta v$ is supervised by VLM logit changes to impose semantic control onto the video generation. Latents $\mathbf{z}'^-_{\Xi},\mathbf{z}'^+_{\Xi}$ obtained from (\ref{eq:gen_pre}), are transported to $\tilde{\mathbf{z}}'^-_\Omega\!\! =\! \Phi(\mathbf{z}'^-_{\Xi},\pi_\Xi^-)$ and $\tilde{\mathbf{z}}'^+_\Omega \!\! = \! \Phi(\mathbf{z}'^+_{\Xi},\pi_\Xi^+)$ within VLM's $\mathbb{R}^\Omega$ space. Respectively, logits $\omega^{-}\!\!=\!\texttt{logit}(\pi_\Omega^-)$ and $\omega^{+}\!\!=\!\texttt{logit}(\pi_\Omega^+))$ are obtained from $\pi_{\Omega}^{-}\!\!=\!\text{LLM}(\beta,\tilde{\mathbf{z}}'^-_\Omega)$ and $\pi_{\Omega}^{+}\!\!=\!\text{LLM}(\beta,\tilde{\mathbf{z}}'^+_\Omega)$ respectively. The logit difference $\Delta \omega$ is computed from their Hellinger distance $\Delta \omega \! =\! \frac{1}{\sqrt{2}} \left| \sqrt{\omega^-} \!-\! \sqrt{\omega^+} \right|$
where $\Delta \omega \in [0,1]$. The obtained divergence is used in (\ref{eq:bank}) as the control parameter $\delta=\Delta \omega$. In training, $\Delta \omega$ is averaged over multiple noise initializations to reduce variance across representations of the same concept/attribute pairs $\pi^-,\pi^+$.
\blfootnote{\textcolor{red}{$^\text{v}$} In house conversion of image method/metric to video.}%

\subsection{From logits to videos}
\label{sec:method::logits}

Inference (\cref{fig:transporter_main_d}) allows $\delta$ to be manually defined to generate videos that reflect specific changes in VLM logit distributions $\Delta \omega$. Using $\mathbf{q}_o$, the approximated probability path (\ref{eq:gen_pre}) is adjusted with condition $\pi^-_\Xi+\delta\mathbf{q}_o$. Generated latents $\mathbf{z}'^{\mathbf{q}_o}_{\Xi,t}$ are decoded to video $\mathbf{x}'^{\mathbf{q}_o}=\mathcal{D}(\mathbf{z}'^{\mathbf{q}_o}_{\Xi,t})$ as in~\cref{fig:deltaomega_bowling}. Additionally, the coupling also enables generating captions for $\mathbf{x}'^{\mathbf{q}_o}$ by transporting $\mathbf{z}'^{\mathbf{q}_o}_{\Xi,t}$ to $\mathbb{R}^{\Omega}$ via $\Phi$.
\section{Experiments}
\label{sec:experiments}

VLMs, training datasets, and \serpantinelight{TRANSPORTER} details are described in~\cref{sec:experiments::details}. L2V is quantitatively compared to adjacent feature visualization tasks over video quality and semantic alignment metrics in~\cref{sec:experiments::results}.  Qualitative results in~\cref{sec:experiments::examples} are followed by
ablations in~\cref{sec:experiments::ablations}.

\subsection{Implementation details}
\label{sec:experiments::details}

\noindent
\textbf{Models and Datasets}. Modulation videos are generated for VideoLLaMA 3 (7B)~\cite{zhang2025videollama}, Gemma 3 (12B)~\cite{team2025gemma}, and Phi 4 MM (5B)~\cite{abouelenin2025phi} logits. The VLM selection is based on ViT/LLM diversity. Wan2.2~\cite{wan2025wan} is used as the base generator. The imported models only run inference with frozen parameters. Coupling is trained on an ensemble of semantically-rich, high-resolution, and egocentric videos from VATEX~\cite{wang2019vatex}, LAVIB~\cite{stergiou2024lavib}, and Ego4D~\cite{grauman2022ego4d}. The concept bank is trained on different seeds of target concepts. 

\noindent
\serpantinemedium{TRANSPORTER} \textbf{settings}. Training is done in two stages. The coupling network $\Phi$ is trained first. $\Phi_{\Omega_1}$ consists of a 24-layer MLPMixer~\cite{tolstikhin2021mlp} projector and $\Phi_{\Omega_2}$ is a 12-layer Transformer~\cite{dosovitskiy2021image}. Module $\rho$-OT uses 100 projections. Coupling is trained with AdamW~\cite{loshchilov2019decoupled} for 100K iterations with a learning rate of $1e^{\!\!-3}$, batch size of 8, and gradients accumulated every 8 steps. Concept bank $\mathbf{Q}$ is trained at a second stage for 1K iterations per vector with $1e^{\!\!-4}$ learning rate. Per generative step, $\epsilon$ is of $16 \!\! \times \!\! 90^2$ size then decoded to $61 \!\! \times \!\! 720^2$ video resolution. $\Delta v$ and $\Delta \omega$ are averaged over multiple seeds. Further details are available in \S \ref{sec:additional_imp_details}.

\subsection{Results}
\label{sec:experiments::results}

\cref{tab:method_comparisons} reports alignment scores between encoded condition text and generated video embeddings, alongside visual quality metrics between generated and real videos. Text-to-video alignment is evaluated with frame-averaged CLIP$^{\textcolor{red}{\text{v}}}$ scores~\cite{hessel2021clipscore}, aesthetic correspondence~\cite{hentschel2022clip}, and cossim divergence score~\cite{baumann2025continuous} between generated $\mathbf{x}'^{\mathbf{q}_o}$ and tokenized target captions $\pi^+$. For video-to-video visual quality, Fr\'echet Video Distance (FVD)~\cite{unterthiner2019fvd} and frame-averaged Learned Perceptual Image Patch Similarity~\cite{zhang2018unreasonable} ($\text{LPIPS}^{\textcolor{red}{\text{v}}}$) are reported between generated and 1K VidChapters7M~\cite{yang2023vidchapters} video embeddings. 
As L2V focuses on visual explanations, baselines include prominent methods from image-~\cite{simonyan2013deep,hatamizadeh2022gradvit,fel2023unlocking} and video-~\cite{stergiou2023leaping} classification, adjusted to maximize a target logit $\omega^+$ or alternatively, logits with varying divergence $\omega^- + \Delta \omega$. Only one previous method has directly addressed visual explanations for video models~\cite{stergiou2023leaping}. Based on its relevance, it is selected as the baseline method for comparisons across settings.

\begin{table}[t]
    \caption{\textbf{Embedding similarity between generated and real videos}. Cosine similarity ($cos$), $l1$/$l2$ distance, and Kullback–Leibler divergence (KL) metrics are compared between mean encodings from VidChapters7M and generated videos. The respective video encoder per VLM is used for each metric.}
    \centering
    \resizebox{\linewidth}{!}{%
    \setlength\tabcolsep{4.0pt}
    \begin{tabular}{l cccc}
        \toprule
         \multirow{2}{*}{Method} & \multicolumn{4}{c}{Metric} \bstrut \\ \cline{2-5}
         & $cos \uparrow$ & $l1\downarrow$ & $l2\downarrow$ & $\text{KL}\downarrow$\tstrut \\
         \multicolumn{5}{l}{\centerrule{2cm} VideoLLaMA 3 \centerrule{3.7cm}} \bstrut \\
         Baseline~\cite{stergiou2023leaping} & 
         $6.45\text{e}\!^{-8}$ & 
         $2.93\text{e}\!^{+2}$ & 
         $2.75\text{e}\!^{+2}$ & 
         56.31 \\
         \serpantinelight{TRANSPORTER}$\;$ & 
         $\mathbf{3.28\textbf{e}\!^{-2}}$ & 
         $\mathbf{5.63\textbf{e}^{0}}$ & 
         $\mathbf{5.88\textbf{e}^{0}}$ & 
         $\mathbf{1.67}$ 
         \\
         \multicolumn{5}{l}{\centerrule{2cm} Gemma 3 \centerrule{4.6cm}} \bstrut \\
         Baseline~\cite{stergiou2023leaping} & $2.31\text{e}\!^{-8}$ & 
         $4.29\text{e}\!^{+2}$ & 
         $5.65\text{e}\!^{+2}$ & 
         $73.00$  
         \\
         \serpantinelight{TRANSPORTER}$\;$ & 
         $\mathbf{1.58\textbf{e}^{-2}}$ & 
         $\mathbf{4.79\textbf{e}^{0}}$ & 
         $\mathbf{7.85\textbf{e}^{0}}$ & 
         $\mathbf{4.00}$ 
         \\
         \multicolumn{5}{l}{\centerrule{2cm} Phi 4 MM \centerrule{4.5cm}} \bstrut \\
         Baseline~\cite{stergiou2023leaping} & 
         $1.31\text{e}\!^{-8}$ &
         $2.47\text{e}\!^{+2}$ & 
         $6.09\text{e}\!^{+2}$ &
         $45.39$  
         \\
         \serpantinelight{TRANSPORTER}$\;$ & 
         $\mathbf{1.25\textbf{e}\!^{-2}}$ & 
         $\mathbf{8.37\textbf{e}^{0}}$ & 
         $\mathbf{1.21\textbf{e}^{0}}$ & 
         $\mathbf{3.66}$ 
         \\
         \bottomrule
    \end{tabular}
    }
    \label{tab:similarity_measures}
    \vspace{-1em}
\end{table}

\noindent
\textbf{Baselines}. Compared to VLM-adjusted prior works, \serpantinelight{TRANSPORTER} yields consistently better semantic alignment and visual quality scores. Notably, as shown in~\cref{tab:method_comparisons::videollama3}, CLIP$^{\textcolor{red}{\text{v}}}$ scores on VideoLLaMA 3 embeddings are improved two-fold with L2V to 35.44, compared to the best AM optimization setting with 16.74. Similarly, significant CLIP$^{\textcolor{red}{\text{v}}}$ aesthetic score improvements are observed. The largest gains reported are for generations conditioned on Phi 4 MM logits, as in~\cref{tab:method_comparisons::phi4}, alongside better embedding cossim divergence score $\Delta$. The suitability of L2V for video model explanations is further evident across visual quality metrics. Videos corresponding to Gemma 3 logits, in~\cref{tab:method_comparisons::gemma3}, are significantly closer to real videos with a $-2.31e^3$ decrease in FVD. Similar trends are also observed for VideoLLaMA 3 logits with $\text{LPIPS}^{\textcolor{red}{\text{v}}}$ consistently reduced by $-5.29$ compared to $\omega^+$ and $\omega^-+\Delta \omega$ AM.

\begin{figure}[t]
    \centering
    \begin{overpic}[width=0.5\textwidth]{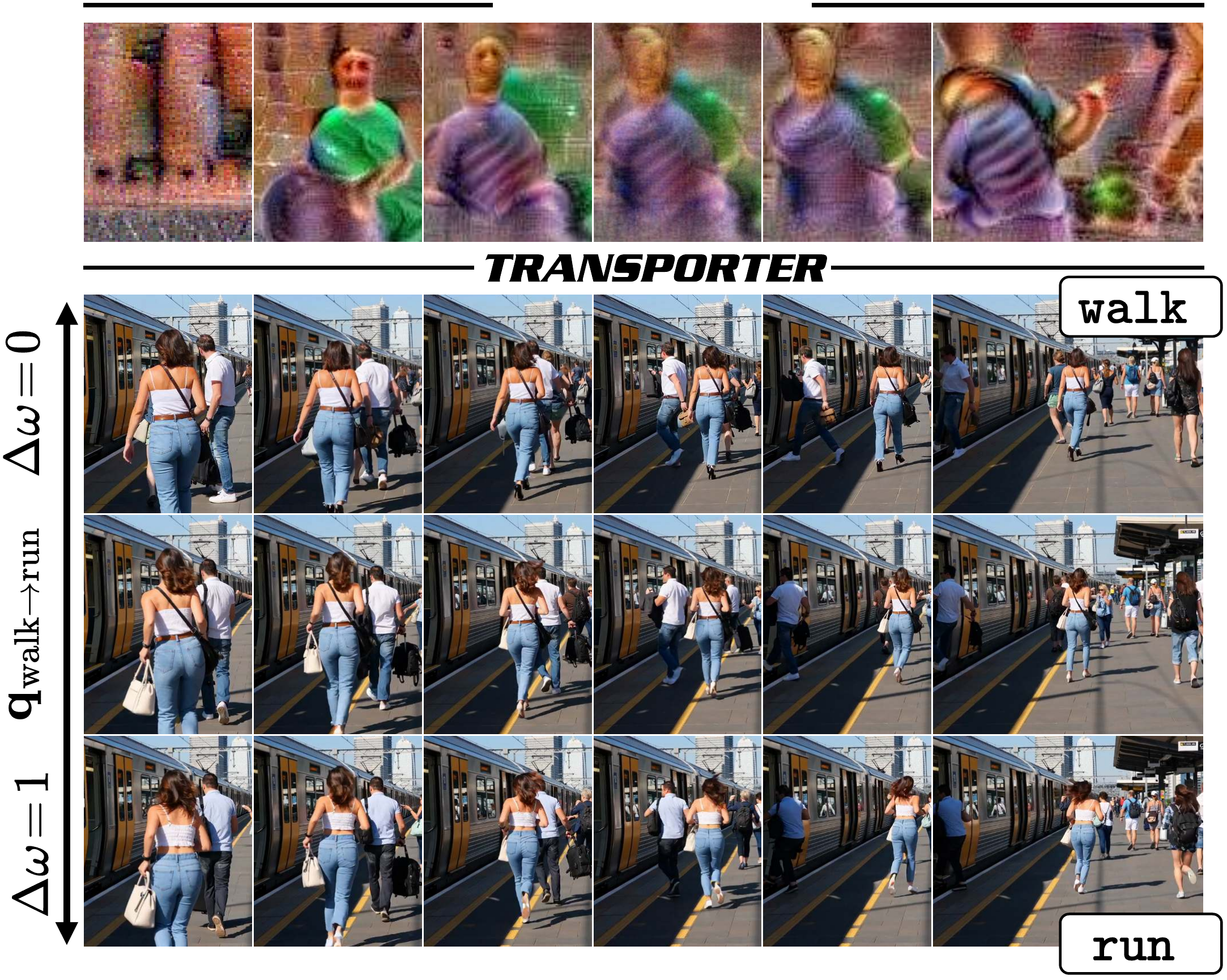}
     \put(43.1,79){Baseline~\cite{stergiou2023leaping}}
     \end{overpic}
    \caption{\textbf{Preferred input generation with AM (top) and proposed L2V (bottom)} based on VideoLLaMA 3 logits corresponding to \inlinebox{\texttt{walk}}. Beyond visualizing single logits, \serpantinelight{TRANSPORTER} further enables generating videos to explore intermediate modulations of the logit distribution when shifting towards \inlinebox{\texttt{run}}.}
    \label{fig:method_comparisons}
    \vspace{-1.2em}
\end{figure}

\begin{figure*}[t]
    \centering
    \includegraphics[width=\linewidth]{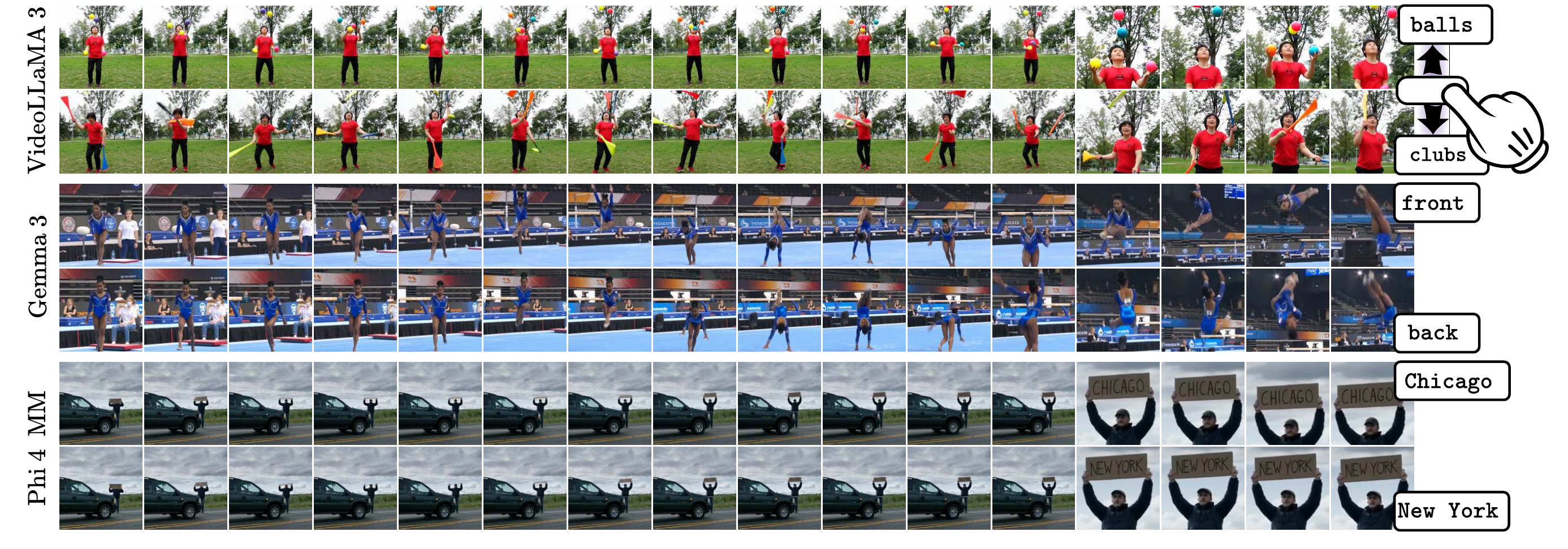}
    \vspace{-0.7em}
    \caption{\textbf{Generated video modulations with }\serpantinemedium{TRANSPORTER}\textbf{across VLMs}. Concept vectors can visualize videos corresponding to logit distributions over a variety of video attributes which can relate to (\textbf{top}) active objects and affordances, such as juggling \inlinebox{\texttt{balls}} or \inlinebox{\texttt{clubs}}, (\textbf{middle}) changes or details in the performance of actions, with \inlinebox{\texttt{front}} and \inlinebox{\texttt{back}} handspring, and (\textbf{bottom}) fine-grained scene details, such as holding a sign that reads \inlinebox{\texttt{Chicago}} or \inlinebox{\texttt{New York}}.}
    \label{fig:qualitative_vlms}
    \vspace{-1em}
\end{figure*}

\begin{figure*}[t]
    \includegraphics[width=\textwidth]{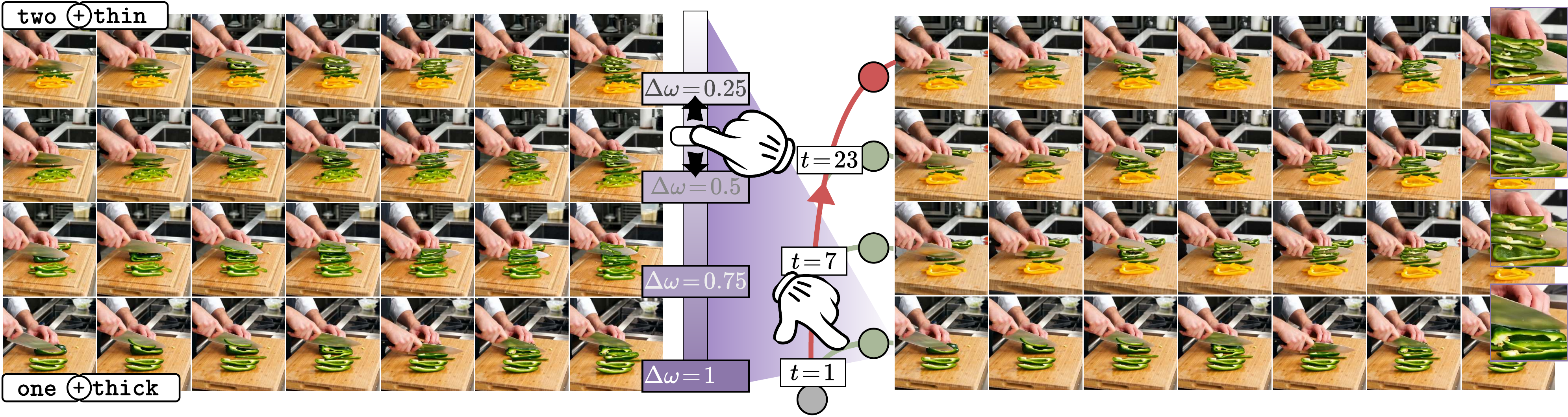}\vspace*{-1.5em}
    \subfloat[\label{fig:ablations_qualitative::a} Logit divergence $\Delta\omega$ modulation for $t=1$]{\hspace{.5\linewidth}}
    \subfloat[\label{fig:ablations_qualitative::b} Step $t$ modulations for for $\Delta \omega=1$.]{\hspace{.5\linewidth}}\vspace{.2em}
    
    \includegraphics[width=\textwidth]{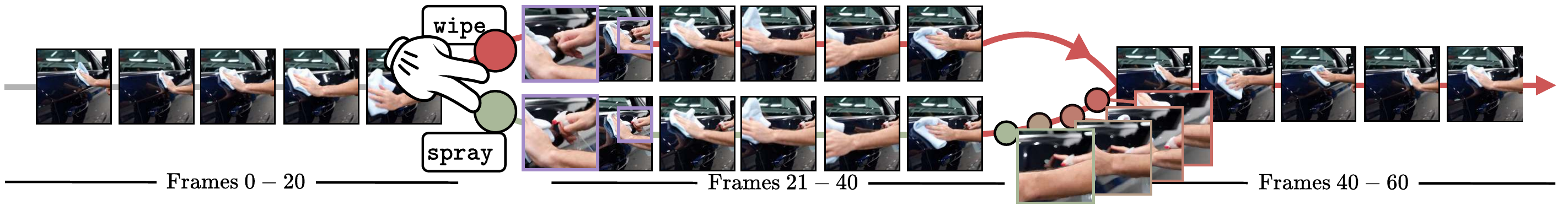}\vspace*{-1em}
    \subfloat[\label{fig:ablations_qualitative::c} Partial latent modulation.]{\hspace{\linewidth}}
    
    \caption{\textbf{Generated videos with Phi 4 MM logits over alternative settings}. (a) \textbf{Divergence modulations} can be done over combined attributes, such as cutting \inlinebox{\texttt{two}} peppers over \inlinebox{\texttt{thin}} strips ($\Delta \omega = 0$) to cutting \inlinebox{\texttt{one}} pepper over \inlinebox{\texttt{thick}} strips ($\Delta \omega = 1$). (b) \serpantinemedium{TRANSPORTER} \textbf{modulations can be introduced at different generation steps} to highlight differences of attribute modulations ($\Delta \omega$) with larger divergence, as that of \inlinebox{\texttt{thin}} and \inlinebox{\texttt{thick}}, being better visible within the first few steps of video generation, compared to smaller concept divergences, such as \inlinebox{\texttt{two}} and \inlinebox{\texttt{one}}. (c) \textbf{Modulations can also be applied partially to latents}. The resulting videos include both original attributes, e.g. \inlinebox{\texttt{wipe}} a car door, and target changes; e.g. \inlinebox{\texttt{spray}} for their respective frames.}
    \label{fig:ablations_qualitative}
    \vspace{-1em}
\end{figure*}

\noindent
\textbf{Multi-metric results}. \cref{tab:similarity_measures} compares embeddings of generated and real videos across VLM encoders. As shown, the baseline struggles to approximate the distribution of real videos. AM does not suffice to capture the high embedding complexity, nor can it effectively disentangle the multi-modal representations of VLMs. With the refined L2V task for visual explanations, \serpantinelight{TRANSPORTER} can generate detailed videos that best correspond to expected token distributions, as shown by the improvements in the cosine similarity and KL divergence scores. In tandem, the videos generated by \serpantinelight{TRANSPORTER} maintain high semantic and visual quality, as evidenced by decreases in distance metrics.

\subsection{Examples}
\label{sec:experiments::examples}

\noindent
\textbf{Qualitative comparisons}. \cref{fig:method_comparisons} illustrates videos generated with different methods over target token \inlinebox{\texttt{walk}}. The AM baseline cannot generate visually distinct frames, resulting in a video containing only an abstract shape. Instead, L2V produces substantially detailed videos of full scenes that capture target attributes. \serpantinelight{TRANSPORTER} is also not limited to representations of individual concepts. The inferred videos can visualize details across attributes, such as expected modulation in pace for tokens corresponding to \inlinebox{\texttt{walk}} or \inlinebox{\texttt{run}}. Effectively, \serpantinelight{TRANSPORTER} is a generative tool that enables users to visualize \textit{what} VLMs' token predictions correspond to.

\begin{table*}[t]
    \caption{\textbf{Semantic scores over} \serpantinemedium{TRANSPORTER} \textbf{ablations} between target caption $\pi^+$ and generated video caption $\pi^{\mathbf{q}_o}$. For each architectural and optimization setting, cosine similarity ($\nabla\!\text{cos}$), BLEU (B@1, B@2, B@3, B@4), CIDEr (C), METEOR (M), and SPICE (S) scores are reported between VideoLLaMA 3/Gemma 3 captions from  \serpantinelight{TRANSPORTER} videos and target captions. Settings are grouped in relation to changes in either the coupling network or the concept bank. Best results are \textbf{bold} and second best are \underline{underlined}.}
    \centering
    \resizebox{\linewidth}{!}{%
    \setlength\tabcolsep{1.8pt}
    \begin{tabular}{ll cccccccc c cccccccc}
        \toprule
         \multicolumn{2}{l}{\multirow{2}{*}{Method}} & \multicolumn{8}{c}{VideoLLaMA 3} & \multicolumn{8}{c}{Gemma 3} \\ 
         \cmidrule(lr){3-10} \cmidrule(lr){12-19}
         && $\nabla\!\cos$  & B@1 & B@2 & B@3 & B@4 & C & M & S & $\;$ & $\nabla\!\cos$ & B@1 & B@2 & B@3 & B@4 & C & M & S \\
        \midrule
        \multicolumn{2}{l}{Baseline~\cite{stergiou2023leaping}} & 0.11 & 20.14 & 10.57 & 5.23 & 2.05 & 2.74 & 11.47 & 9.12 && 0.05 & 20.62 & 11.18 & 5.73 & 2.19 & 1.95 & 11.60 & 7.55 \\
        \addlinespace[0.1em]
        \multicolumn{2}{l}{\centerrule{3cm}} &\multicolumn{17}{l}{Coupling network ablations \centerrule{10.67cm}} \bstrut \\
        \multicolumn{2}{l}{decode-and-re-encode} & 0.19 & 28.86 & 16.49 & 10.38 & 5.24 & 22.72 & 14.94 & 23.45 && 0.11 & 29.36 & 17.80 & 11.07 & 8.77 & 15.36 & 16.23 & 25.03 \\
        \multicolumn{2}{l}{$\Phi_{\Omega_1}$ only} & 0.16 & 26.75 & 15.79 & 8.67 & 4.49 & 18.17 & 13.28 & 18.78 && 0.09 & 28.02 & 15.59 & 10.39 & 7.46 & 14.11 & 15.92 & 23.58 \\
        \multicolumn{2}{l}{$\Phi_{\Omega_2}$ only} & 0.21 & 30.48 & 18.20 & 11.73 & 7.75 & 25.56 & 15.09 & 25.21 && 0.13 & 31.69 & 19.43 & 13.64 & 10.57 & 18.36 & 20.92 & 26.57 \\
        \multicolumn{2}{l}{$\text{mean}\langle\Phi_{\Omega_1,\Omega_2}\rangle$} & 0.23 & 31.25 & 19.39 & 11.99 & 7.79 & 26.70 & 17.16 & 25.46 && 0.14 & 33.21 & 21.57 & 15.14 & 11.21 & 19.88 & 21.42 & 27.20 \\
        \multicolumn{2}{l}{sGW OT} & 0.14 & 24.87 & 13.43 & 6.71 & 4.09 & 7.92 & 12.46 & 11.88 && 0.08 & 27.47 & 13.74 & 7.56 & 4.41 & 7.00 & 13.87 & 9.24 \\
        \multicolumn{1}{r}{\multirow{3}{*}{$P\Bigg\{$}} & \multicolumn{1}{l}{$50$} & \underline{0.24} & 34.04 & 20.18 & 12.36 & 8.29 & 27.63 & 20.12 & 29.27 && 0.13 & 35.45 & 22.16 & 15.24 & 11.64 & 20.87 & 22.12 & 28.89 \\
         & \multicolumn{1}{l}{$200$} & \textbf{0.26} & \underline{34.63} & \underline{20.64} & \underline{12.69} & 8.30 & \textbf{28.14} & \underline{20.34} & \underline{29.62} && \underline{0.15} & \underline{35.84} & \underline{22.65} & 15.59 & \textbf{11.80} & 21.30 & \underline{22.57} & \underline{29.88} \\
         & \multicolumn{1}{l}{$400$} & \textbf{0.26} & \textbf{34.72} & \textbf{20.71} & \textbf{12.82} & \textbf{8.40} & \underline{28.13} & \textbf{20.41} & \textbf{29.65} && \textbf{0.16} & \textbf{35.93} & \underline{22.65} & \textbf{15.70} & \textbf{11.80} & \underline{21.31} & \textbf{22.63} & \textbf{29.96} \\
        \addlinespace[0.1em]
        \multicolumn{2}{l}{\centerrule{3cm}} &\multicolumn{17}{l}{Concept bank ablations \centerrule{11.25cm}} \bstrut \\
        \multirow{2}{*}{$\Delta\omega\biggl\{$} & KL & 0.19 & 29.83 & 17.56 & 10.45 & 5.29 & 24.37 & 15.72 & 24.80 && 0.10 & 30.56 & 18.25 & 11.44 & 9.69 & 17.09 & 19.34 & 26.52 \\
         & JS & 0.22 & 32.48 & 19.07 & 11.36 & 6.58 & 26.26 & 18.93 & 26.75 && 0.13 & 33.13 & 19.88 & 13.04 & 10.61 & 19.78 & 21.33 & 27.76 \\
        \midrule
        \multicolumn{2}{l}{\serpantinelight{TRANSPORTER}} & \textbf{0.26} & 34.59 & 20.59 & 12.67 & \underline{8.34} & 27.99 & \underline{20.34} & 29.59 && \underline{0.15} & 35.80 & \textbf{22.70} & \underline{15.65} & \underline{11.78} & \textbf{21.33} & 22.51 & 29.85 \\
        \bottomrule
    \end{tabular}
    }
    \label{tab:transporter_ablations}
    \vspace{-1em}
\end{table*}

\noindent
\textbf{Generalizability}. \cref{fig:qualitative_vlms} demonstrates \serpantinelight{TRANSPORTER}'s applicability to a range of VLMs and its capability to generate videos that maintain scene aesthetics. As shown in the obtained videos, VLMs reason about distinct attributes, such as the types of objects used in actions, \textit{e.g.} juggling, or what a sign reads, throughout entire scenes. However, changes in action performance, \textit{e.g.}, gymnastics spin, seem to be more influential at specific times, without necessarily having long-term effects. This shows that current models learn action sequentiality over different granularities.  

\noindent
\textbf{Logit divergence}. \cref{fig:ablations_qualitative::a} shows videos generated over multiple attribute modulations across logit divergences. Interestingly, general attributes such as the number of objects, \inlinebox{\texttt{two}} or \inlinebox{\texttt{one}}, exhibit greater variance in the logit divergence, as shown by multiple $\Delta \omega$ values. In comparison, finer details such as object interactions with \inlinebox{\texttt{thin}} or \inlinebox{\texttt{thick}} cuts are only present for a fraction of $\Delta \omega$. This shows that learned semantic correspondences are, in turn, reflected onto token divergences.   

\noindent
\textbf{Flow timestep selection}. The effects of modulations over different timesteps are shown in~\cref{fig:ablations_qualitative::b}. Modulations across large distributions are generated first, as evidenced by the immediate change in the number of peppers. Instead, later generative steps focus on finer details, such as the slicing thickness, which become progressively less visible as modulations are added in subsequent generation steps.

\noindent
\textbf{Partial conditioning}.~\cref{fig:ablations_qualitative::c} presents videos with only a fraction of their frames modulated during generation. The visualizations demonstrate how and at what intensity target changes are expected to occur within time segments as actions unfold. As shown, a spray bottle blends into the \inlinebox{\texttt{wipe}} to \inlinebox{\texttt{spray}} transition, showing how VLMs bind objects to actions and to adjacent actions.

\subsection{Ablations}
\label{sec:experiments::ablations}

Ablations are performed for both optimization steps and method settings.
    
\noindent
\textbf{Coupling network settings}. To investigate the impact of each coupling network component, results between condition captions and generated LLM captions from \serpantinelight{TRANSPORTER} videos are reported at the top of~\cref{tab:transporter_ablations}. Notably, a significant improvement across all semantic metrics is observed with the coupling modules $\Phi_{\Omega_1}/\Phi_{\Omega_2}$ compared to decoding and re-encoding the generated latents, as in the inference setting. For OT, the heuristic approach based on Sliced Gromov-Wasserstein (sGW)~\cite{titouan2019sliced} shows a noticeable decrease in caption quality. Increasing the number of projections offers some marginal improvements, but introduces additional computational costs as discussed in \S \ref{sec:heuristic_ot}. 

\noindent
\textbf{Concept bank divergence metric}. The second half of~\cref{tab:transporter_ablations} ablates divergence metrics. As shown, normalizing the bounds of the divergence metrics affects performance. Hillinger distance is chosen given its strict bounds and slightly favorable performance to the Jaccard Similarity (JS). Unbounded metrics, such as the KL divergence, resulted in lower performance.

\begin{figure}
    \centering
        \includegraphics[width=\linewidth]{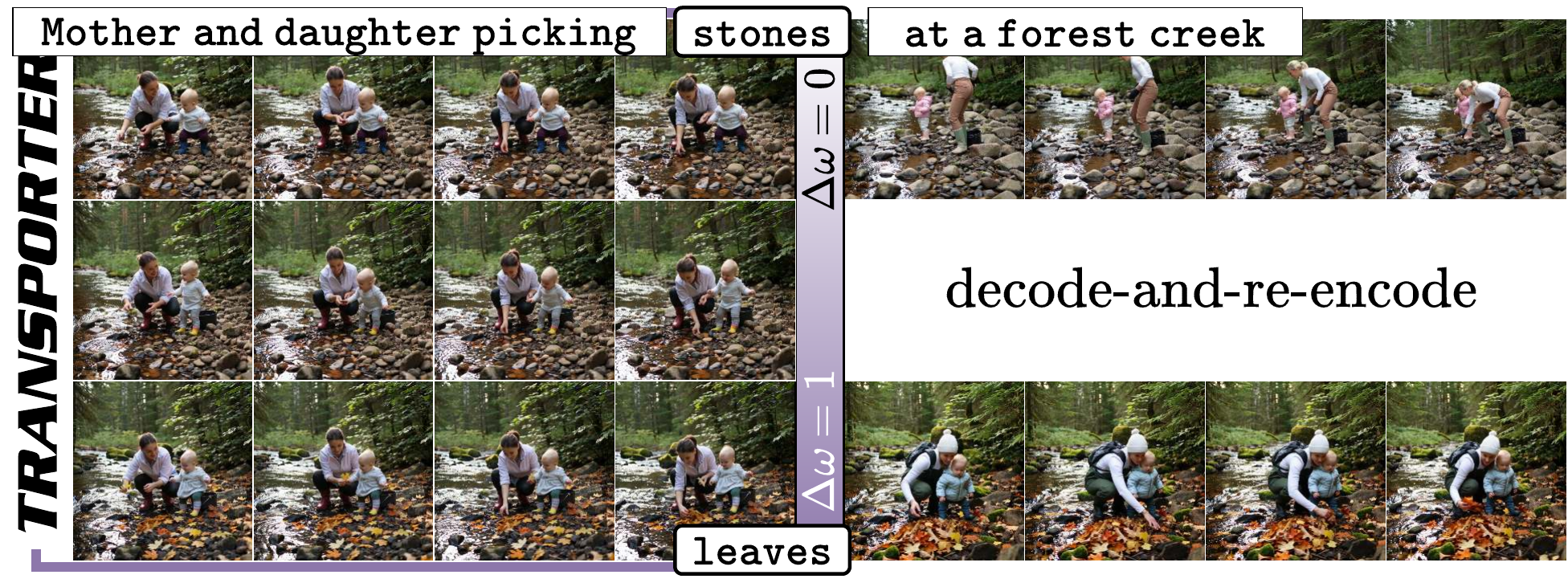}
        \vspace{-1.8em}
        \captionof{figure}{\serpantinemedium{TRANSPORTER}\textbf{comparison to decode-and-re-encode inference} for Gemma 3 logit divergence between \inlinebox{\texttt{stone}} and \inlinebox{\texttt{leaves}}. Learned modulations with \serpantinelight{TRANSPORTER} maintain scene aspects unchanged. This robustness does not translate in inference-only settings.}
        \label{fig:ablations:baseline}
        \vspace{-1em}
\end{figure}

\noindent
\textbf{Inference only}. \Cref{fig:ablations:baseline} presents qualitative comparisons to decode-and-re-encode setting. \serpantinelight{TRANSPORTER} reflects VLM token modulations across different logit differences, which is not possible if only using models at inference. The visualizations further demonstrate \serpantinelight{TRANSPORTER}'s ability to maintain the motion and dynamics of the original scene while visually reflecting the target changes across relevant VLM tokens.

\section{Conclusion}
\label{sec:conclusion}

This paper introduces logits-to-video (L2V). A generative task for video model explainability that goes beyond the adaptation of current classifier-based approaches to video VLMs. Alongside L2V, the paper proposes \serpantinelight{TRANSPORTER}, a visual representation method for VLM token predictions. During training, \serpantinelight{TRANSPORTER} learns the optimal transport between semantic VLM spaces and the visual spaces of generator models. From this coupling, the divergence between tokens is used to define modulations in the embedding space corresponding to attributes. During inference, \serpantinelight{TRANSPORTER} generates videos to represent this divergence.  The proposed method has shown qualitatively
and quantitatively that the explanations generated are of high visual quality and semantically aligned to target modulations. The video fidelity alongside the high visual and semantic scores make L2V and \serpantinelight{TRANSPORTER} a first step towards visualizing VLMs' predictions and a novel direction for understanding their reasoning process.

\section*{Acknowledgments}
Research used the Dutch national e-infrastructure supported by the
SURF Cooperative with grant EINF-15225.

{
    \small
    \bibliographystyle{ieeenat_fullname}
    \bibliography{main}
}

\clearpage
\setcounter{page}{1}
\maketitlesupplementary

\begin{algorithm}[H]
\caption{$\rho$-OT}
\label{alg:plot}
\textbf{Input}:\\
\hspace*{\algorithmicindent} Embeddings $\widehat{\textbf{z}}_{\Omega_1}$, $\widehat{\textbf{z}}_{\Omega_2} $ \\
\hspace*{\algorithmicindent} Projection vectors $\{\mathbf{p}_{\Omega_1,\rho}\}_{\rho=1}^P$, $\{\mathbf{p}_{\Omega_2,\rho}\}_{\rho=1}^P$ \\
\hspace*{\algorithmicindent} Temp. $\tau$, Reg. strength $\lambda$, Sinkhorn iter. $K$ \\
\hspace*{\algorithmicindent} Uniform $|\mathbf{N}| \times |\mathbf{N}|$ matrix $\mathbf{U}$ (where $\mathbf{U}_{i,j} = 1/|\mathbf{N}|$) \\
\textbf{Output}:\\
\hspace*{\algorithmicindent} Transported $\tilde{\mathbf{z}}_\Omega = \tilde{\mathbf{T}} \widehat{\mathbf{z}}_{\Omega_1}$ 
\begin{algorithmic}[1]
\State $\mathbf{T} \leftarrow \mathbf{0} \in \mathbb{R}^{|\mathbf{N}| \times |\mathbf{N}|}$
\For{$\rho \in \{1,\dots,P\}$} 
    \State $\mathbf{a}_{\rho} \leftarrow \langle \widehat{\mathbf{z}}_{\Omega_1,i},\mathbf{p}_{\Omega_1,\rho} \rangle$
    \State $\mathbf{b}_{\rho} \leftarrow \langle \widehat{\mathbf{z}}_{\Omega_2,j},\mathbf{p}_{\Omega_2,\rho} \rangle$
    
    \State $\mathbf{M}_{\rho} \leftarrow \langle -\|\mathbf{a}_{i,\rho} - \mathbf{b}_{j,\rho}\|_2 / \tau \rangle$ \Comment{Neg. dist}

    \State $\mathbf{M}_{\text{max}} \leftarrow \langle \text{max}_j(\mathbf{M}_{\rho,i,j})\rangle$
    \State $\mathbf{R}_{\rho} \leftarrow \langle \text{exp}(\mathbf{M}_{\rho,i,j} - \mathbf{M}_{\text{max},i}) \rangle$
    
    \State $\mathbf{T}_{\rho} \leftarrow \text{diag}(\mathbf{R}_{\rho} \mathbf{1})^{-1} \mathbf{R}_{\rho}$ \Comment{Row softmax}
    
    \State $\mathbf{T}_{\rho,\lambda} \leftarrow (1-\lambda)\mathbf{T}_{\rho} + \lambda\mathbf{U}$ \Comment{Uniform reg}
    
    \State $\mathbf{T} \leftarrow \mathbf{T} + \mathbf{T}_{\rho,\lambda}$
\EndFor
\State $\tilde{\mathbf{T}} \leftarrow \mathbf{T} / P$ \Comment{Avg slice plans}

\For{$k \in \{1,\dots,K\}$} \Comment{Sinkhorn-Knopp}
    \State $\tilde{\mathbf{T}} \leftarrow \text{diag}(\tilde{\mathbf{T}} \mathbf{1})^{-1} \tilde{\mathbf{T}}$ \Comment{Row norm}
    \State $\tilde{\mathbf{T}} \leftarrow \tilde{\mathbf{T}} \text{diag}(\mathbf{1}^T \tilde{\mathbf{T}})^{-1}$ \Comment{Column norm}
\EndFor

\State \Return $\tilde{\mathbf{T}} \widehat{\mathbf{z}}_{\Omega_1}$
\end{algorithmic}
\end{algorithm}

\section{Entropic Optimal Transport Formulation}
\label{sec:entropic_ot}

Transport plan $\gamma_\rho$ is found by minimizing transport cost $\mathbf{M}\!=\!|\mathbf{a}\!-\!\mathbf{b}|$ and maximizing entropy:
\begin{equation}
    \underset{\gamma_\rho}{\text{min}}\!\underbrace{\int_{\mathbb{R} \times \mathbb{R}} \!\!\!\!\!\!\!\! \mathbf{M} d\gamma_\rho(\mathbf{a},\mathbf{b})}_{\text{Transport cost}} \! - \! \tau \!\! \underbrace{\int_{\mathbb{R} \times \mathbb{R}} \!\!\!\!\!\!\!\! \gamma_\rho(\mathbf{a},\mathbf{b})\text{log}(\gamma_\rho(\mathbf{a},\mathbf{b}))d\mathbf{a}d\mathbf{b}}_{\text{Shannon entropy}}
\label{eq:eOT}
\end{equation}
\noindent
where the differential $d\gamma_\rho(\mathbf{a},\mathbf{b})$ is the probability mass from $\mathbf{a} \! \rightarrow \! \mathbf{b}$ alogside entropic regularization. A discrete form of (\ref{eq:eOT}) for $\mathbf{N}$ tokens and transport plan $\mathbf{T}_\rho \! \in \! \mathbb{R}^{|\mathbf{N}| \times |\mathbf{N}|}$ over cost matrix $\mathbf{M}_{i,j,\rho}\! =\! \|\mathbf{a}_{i,\rho}\!-\!\mathbf{b}_{j,\rho}\|_2$, given temperature $\tau$, can be formulated to find optimal transport $\tilde{\mathbf{T}}$:
\begin{equation}
    \tilde{\mathbf{T}} \!\! = \!\! \underset{\mathbf{T}_\rho}{\text{argmin}}\frac{1}{P}\!\!\sum_{\rho=1}^{P}\Bigl(\!\underbrace{\sum_{i,j \in \mathbf{N}}\! \underset{i,j,\rho}{\mathbf{T}} \;\underset{i,j,\rho}{\mathbf{M}}}_{\text{Transport cost}} \!\! +\! \tau \!\!  \underbrace{ \sum_{i,j\in\mathbf{N}}\!\!\underset{i,j,\rho}{\mathbf{T}}\!\!\log(\underset{i,j,\rho}{\mathbf{T}})}_{\text{Entropy regularization}}\Bigr)
\label{eq:nOT}
\end{equation}
Solving the \emph{full} doubly-constrained problem in~(\ref{eq:nOT}) is computationally intensive. Thus, the proposed learnable $\rho$-OT provides an efficient approximation.

\section{Procedural details for OT}
\label{sec:rho_ot_formulation}

Projected embeddings $\widehat{\mathbf{z}}_{\Omega_1}$ are learned to minimize their mean divergence to target $\mathbf{z}_{\Omega}$. However, this does not guarantee that target local token structures are learned similar to those of $\widehat{\mathbf{z}}_{\Omega_2}$ optimized with the Gram-matrix loss. To combine their properties, learnable vectors $\mathbf{p}_{\Omega_1},\mathbf{p}_{\Omega_2}$ are used to compute an optimal transport plan as shown in~\cref{alg:plot}. Embeddings $\widehat{\mathbf{z}}_{\Omega_1},\widehat{\mathbf{z}}_{\Omega_2}$ are projected to $\mathbf{a},\mathbf{b}$ based on their inner product with with $\mathbf{p}_{\Omega_1,\rho},\mathbf{p}_{\Omega_2,\rho}$. Cost matrix $\mathbf{M}_\rho$ per $\rho$ is computed from their negative pair-wise l2 distance, scaled by the temperature $\tau$. This avoids computing a single, high-dimensional cost matrix and instead computes multiple, simpler cost matrices. The cost also enforces $\mathbf{p}_{\Omega_1},\mathbf{p}_{\Omega_2}$ to produce projections with strong correspondence across the diagonal $\mathbf{a}_{i,\rho} \approx \mathbf{b}_{i,\rho}$. For numeric stability, the cost matrix $\mathbf{M}_\rho$ is normalized by row-wise softmax for $P$. To avoid sparsity, $\tilde{\mathbf{T}}$ is regularized with a uniform distribution $\frac{1}{|\mathbf{N|}}\mathbf{U}$ with weight $\lambda$. The final transport plan $\tilde{\mathbf{T}}$ is computed by averaging the sliced plans, followed by $K=3$ Sinkhorn-Knopp~\cite{knight2008sinkhorn} iterations to ensure a better approximation to the doubly-stochastic matrix of~(\ref{eq:nOT}).

\begin{table}[t]
\centering
    \caption{\textbf{VLM architecture settings}. VideoLLaMA 3 (V), Gemma 3, (G), and Phi 4 MM (P), use different vision encoders, input resolutions, multi-modal projectors, and backbone LLMs.}
    \resizebox{\linewidth}{!}{%
    \setlength\tabcolsep{1.5pt}
    \begin{tabular}{l lc c lc}
        \toprule
        \multirow{2}{*}{VLM} & \multicolumn{2}{c}{Vision Encoder $\mathcal{E}$} & \multicolumn{1}{c}{MM proj} & \multicolumn{2}{c}{LLM} \\ \cline{2-3} \cline{5-6} 
         & Model & res & dim & Model & prms \\
         \midrule
         V & SoViT-400m/14~\cite{alabdulmohsin2023getting} & 384$^2$ & 3584 & Qwen2.5~\cite{bai2025qwen2} & 7B \\ 
         G & \multirow{2}{*}{SigLIP-so400m/14~\cite{tschannen2025siglip}} & 896$^2$ & 3584 & Gemma 3~\cite{team2025gemma} & 12B \\
         P & & [28$^2$,448$^2$]& 3072 & Phi 4 Mini~\cite{abouelenin2025phi} & 5B\\
         \bottomrule
    \end{tabular}
    }
    \label{tab:vlm_details}
    \vspace{-1em}
\end{table}

\section{Additional implementation details}
\label{sec:additional_imp_details}

\noindent
\textbf{VLM selection}. An overview of the selected VLM details is provided in~\cref{tab:vlm_details}. As shown, the models represent a wide spectrum of design choices varying in fundamental components such as visual encoders, embedding space size, and backbone language models. This heterogeneity aims to better reflect the general applicability of \serpantinelight{TRANSPORTER}, as the presented results are not bound to a single model family or design paradigm, enabling a comprehensive analysis.

\begin{table}[t]
    \caption{\textbf{FVD and CLIPScores across dataset settings}. Evaluation metrics are reported as in~\cref{tab:method_comparisons}. Best results are in \textbf{bold} and the used setting is in \textcolor{cvprblue}{blue}.}
    \centering
    \resizebox{\linewidth}{!}{
    \setlength\tabcolsep{1.8pt}
    \begin{tabular}{cc| cccc}
    \toprule
         &\multicolumn{1}{c}{}& \multicolumn{4}{c}{EGO4D} \bstrut \\ \cline{3-6}
         & \multicolumn{1}{c}{}& 40K & 80K & 120K & 160K \tstrut \\
         \midrule
         \multicolumn{1}{c|}{\multirow{4}{*}{\rotatebox[origin=c]{90}{LAVIB}}} & 40K & 2.12e$^{\!2}$/34.86 & 1.72e$^{\!2}$/35.10 & 1.33e$^{\!2}$/35.41 & 1.23e$^{\!2}$/35.43 \\
         \multicolumn{1}{c|}{} & 80K & 1.65e$^{\!2}$/35.24 & \textcolor{cvprblue}{1.25e$^{\!2}$}/\textcolor{cvprblue}{35.44} & 1.22e$^{\!2}$/35.50 & 1.20e$^{\!2}$/35.52 \\
         \multicolumn{1}{c|}{} & 120K & N/A & 1.23e$^{\!2}$/35.50 & 1.19e$^{\!2}$/35.53 & \textbf{1.18e}$^{\!\mathbf{2}}$/\textbf{35.54} \\
         \multicolumn{1}{c|}{} & 160K & N/A & 1.19e$^{\!2}$/\textbf{35.54} & N/A & N/A\\
         \bottomrule
    \end{tabular}
    }
    \label{tab:dataset_settings}
    \vspace{-1em}
\end{table}

\noindent
\textbf{Coupling network datasets}. In total, 200K videos are used for training the coupling network. This includes 40K VATEX, 80K LAVIB, and 80K Ego4D annotated clips. As VATEX is the smallest of the three, all training videos are used. In contrast, 160K total videos from LAVIB and Ego4D are selected. The choice of 200K iterations and dataset distributions is found by the grid search in~\cref{tab:dataset_settings}. Larger dataset samples only provide marginal improvements over the coupling network's reconstruction task, but with larger computational overheads. Thus, an 80/80K split is selected while taking into account training setup ablations.

\begin{table*}[t]
    \caption{\textbf{Prompts and modulations used for exploring active object, action performance, scene context, and combined settings logits}. Source-to-target modulations across Video LLaMA 3 (v), Gemma 3 (G), and Phi 4 MM (P) logits are denoted between $\pi^-$ and target $\pi^+$.}
    \label{tab:prompts}
    \centering
    \resizebox{\linewidth}{!}{%
    \begin{tabular}{cc lll}
    \toprule
    \multirow{2}{*}{VLM} & \multirow{2}{*}{Fig.} &
    \multirow{2}{*}{prompt (\texttt{A ...})} & \multicolumn{2}{c}{Modulations} \\
    &&& \multicolumn{1}{c}{$\pi^-$} & \multicolumn{1}{c}{$\pi^+$} \\

    \multicolumn{5}{l}{\centerrule{3cm} Active Objects \centerrule{28.3cm}} \bstrut \\

    G&\ref{fig:deltaomega_bowling}&\texttt{close up shot of a} \inlinebox{\textcolor{white}{attr}} \texttt{bowling ball hitting pins in a bowling alley} & \inlinebox{\texttt{red}} & \inlinebox{\texttt{blue}} \\

    V&\ref{fig:qualitative_vlms}&\texttt{person jugging} \inlinebox{\textcolor{white}{attr}} \texttt{ outdoors}. & \inlinebox{\texttt{balls}} & \inlinebox{\texttt{clubs}} \\

    V&\ref{fig:qualitative::object:::pink}&\texttt{person pushing a } \inlinebox{\textcolor{white}{attr}} \texttt{stroller outdoorns during a sunny day.} & \inlinebox{\texttt{purple}} & \inlinebox{\texttt{pink}} \\

    V&\ref{fig:qualitative::object:::starfish}&\texttt{video of a scuba diver swimming underwater in the ocean and picking up a } \inlinebox{\textcolor{white}{attr}}. & \inlinebox{\texttt{starfish}} & \inlinebox{\texttt{seashell}} \\

    V&\ref{fig:qualitative::object:::cookies}&\texttt{close-up of a person baking } \inlinebox{\textcolor{white}{attr}} \texttt{cookies. The cookies are on a baking tray.} & \inlinebox{\texttt{christmas tree}} & \inlinebox{\texttt{circle}} \\

    G&\ref{fig:qualitative::object:::wood}&\texttt{video of a person writing TRANSPORTER on a } \inlinebox{\textcolor{white}{attr}} \texttt{table}. & \inlinebox{\texttt{wood}} & \inlinebox{\texttt{metal}} \\

    G&\ref{fig:qualitative::object:::book}&\texttt{video of a person reading a} \inlinebox{\textcolor{white}{attr}} \texttt{while sitting on a park bench}. & \inlinebox{\texttt{book}} & \inlinebox{\texttt{newspaper}} \\

    G&\ref{fig:qualitative::object:::mango}&\texttt{person picking a(n)} \inlinebox{\textcolor{white}{attr}} \texttt{in a grocery store with shelves of fruits and vegetables in the background.}& \inlinebox{\texttt{mango}} & \inlinebox{\texttt{apple}} \\

    P&\ref{fig:qualitative::object:::roll}&\texttt{person painting a wall with a} \inlinebox{\textcolor{white}{attr}}. & \inlinebox{\texttt{roll}} & \inlinebox{\texttt{brush}} \\

    P&\ref{fig:qualitative::object:::typewriter}&\texttt{top-view video of a person typing on a } \inlinebox{\textcolor{white}{attr}}. & \inlinebox{\texttt{typewriter}} & \inlinebox{\texttt{keyboard}} \\

    \multicolumn{5}{l}{\centerrule{3cm} Action performance \centerrule{27.5cm}} \bstrut \\

    V&\ref{fig:method_comparisons}&\texttt{video of a person} \inlinebox{\textcolor{white}{attr}} \texttt{to get on the train}. & \inlinebox{\texttt{walking}} & \inlinebox{\texttt{running}} \\

    G&\ref{fig:qualitative_vlms}&\texttt{artistic gymnast performing a routine landing a dismount with a} \inlinebox{\textcolor{white}{attr}} \texttt{ handspring and a full twist}. & \inlinebox{\texttt{front}} & \inlinebox{\texttt{back}} \\

    P&\ref{fig:ablations_qualitative::c}&\texttt{close-up video of a person} \inlinebox{\textcolor{white}{attr}} \texttt{ a car door with a cloth}. & \inlinebox{\texttt{wiping}} & \inlinebox{\texttt{spraying}} \\

    V&\ref{fig:qualitative::action:::surf}&\texttt{video of a surfer } \inlinebox{\textcolor{white}{attr}} \texttt{at the open sea.} & \inlinebox{\texttt{surfing}} & \inlinebox{\texttt{kite surfing}} \\

    V&\ref{fig:qualitative::action:::skate}&\texttt{a video of a teenager } \inlinebox{\textcolor{white}{attr}} \texttt{ at a public staircase in a city center}. & \inlinebox{\texttt{roller skating}} & \inlinebox{\texttt{skateboarding}} \\

    V&\ref{fig:qualitative::action:::run}&\texttt{person } \inlinebox{\textcolor{white}{attr}} \texttt{in a lush field full of flowers} & \inlinebox{\texttt{running}} & \inlinebox{\texttt{spinning}} \\

    G&\ref{fig:qualitative::action:::roll}&\texttt{video of a chef } \inlinebox{\textcolor{white}{attr}} \texttt{pizza dough in his kitchen} & \inlinebox{\texttt{rolling}} & \inlinebox{\texttt{stretching}} \\

    G&\ref{fig:qualitative::action:::stir}&\texttt{video of a person that } \inlinebox{\textcolor{white}{attr}}\texttt{s a cup of coffee in the kitchen.} & \inlinebox{\texttt{stir}} & \inlinebox{\texttt{pour}} \\

    G&\ref{fig:qualitative::action:::ride}&\texttt{a cowboy} \inlinebox{\textcolor{white}{attr}} \texttt{ a brown horse in the outback.}. & \inlinebox{\texttt{riding}} & \inlinebox{\texttt{leading}} \\

    P&\ref{fig:qualitative::action:::bounce}&\texttt{person } \inlinebox{\textcolor{white}{attr}} \texttt{a soccer ball at a soccer field.} & \inlinebox{\texttt{toe bouncing}} & \inlinebox{\texttt{head bouncing}} \\

    P&\ref{fig:qualitative::action:::pick}&\texttt{video of a person that} \inlinebox{\textcolor{white}{attr}} \texttt{boxes to his car.} & \inlinebox{\texttt{picks up}} & \inlinebox{\texttt{drags}} \\

    \multicolumn{5}{l}{\centerrule{3cm} Scene context\centerrule{28.6cm}} \bstrut \\

    V&\ref{fig:transporter_main}&\texttt{video of a band playing music on a promenade. The camera zooms to the musician on the} \inlinebox{\textcolor{white}{attr}} \texttt{of the scene}. & \inlinebox{\texttt{left}} & \inlinebox{\texttt{right}} \\

    P&\ref{fig:qualitative_vlms}&\texttt{hitchhiker standing on the side of the road and puts up a cardboard sign that says} \inlinebox{\textcolor{white}{attr}}. & \inlinebox{\texttt{Chicago}} & \inlinebox{\texttt{New York}} \\

    V&\ref{fig:qualitative::scene:::parking}&\texttt{person existing a classic car parked at a} \inlinebox{\textcolor{white}{attr}}. & \inlinebox{\texttt{parking lot}} & \inlinebox{\texttt{underpass}} \\
    V&\ref{fig:qualitative::scene:::sandtrap}&\texttt{golfer hitting the ball while on} \inlinebox{\textcolor{white}{attr}}. & \inlinebox{\texttt{grass}} & \inlinebox{\texttt{sandtrap}} \\

    G&\ref{fig:qualitative::scene:::outdoor}&\texttt{person painting a wall} \inlinebox{\textcolor{white}{attr}}. & \inlinebox{\texttt{outdoors}} & \inlinebox{\texttt{indoors}} \\

    G&\ref{fig:qualitative::scene:::outdoor}&\texttt{video of a tennis player getting ready to serve the ball on a} \inlinebox{\textcolor{white}{attr}} \texttt{court}. & \inlinebox{\texttt{clay}} & \inlinebox{\texttt{grass}} \\

    P&\ref{fig:qualitative::scene:::tree}&\texttt{group of firefighters rescuing a cat stranded on a} \inlinebox{\textcolor{white}{attr}}. & \inlinebox{\texttt{tree}} & \inlinebox{\texttt{roof}} \\

    P&\ref{fig:qualitative::scene:::sky}&\texttt{video showing a group of people playing beach volleyball on a day with} \inlinebox{\textcolor{white}{attr}}. & \inlinebox{\texttt{overcast}} & \inlinebox{\texttt{clear sky}} \\

    \multicolumn{5}{l}{\centerrule{3cm} Combined modulations \centerrule{27cm}} \bstrut \\

    P&\ref{fig:ablations_qualitative::a},\ref{fig:ablations_qualitative::b}&\texttt{chef cutting} \inlinebox{\textcolor{white}{attr}} \texttt{into} \inlinebox{\textcolor{white}{attr}} \texttt{pieces}. & \inlinebox{\texttt{two}},\inlinebox{\texttt{thin}} & 
    \inlinebox{\texttt{one}},\inlinebox{\texttt{thick}} \\

    V&\ref{fig:qualitative::multi:::bike} & \texttt{person riding a} \inlinebox{\textcolor{white}{attr}} \texttt{while wearing a} \inlinebox{\textcolor{white}{attr}} \texttt{and} \inlinebox{\textcolor{white}{attr}} \texttt{jeans}.& \inlinebox{\texttt{bicycle}},\inlinebox{\texttt{jacket}},\inlinebox{\texttt{black}} & \inlinebox{\texttt{scooter}},\inlinebox{\texttt{t-shirt}},\inlinebox{\texttt{blue}} \\

    G&\ref{fig:qualitative::multi:::black}&\texttt{video of a person with} \inlinebox{\textcolor{white}{attr}} \texttt{hair color during a} \inlinebox{\textcolor{white}{attr}} \texttt{walk around the city}. & \inlinebox{\texttt{black}},\inlinebox{\texttt{night}} & \inlinebox{\texttt{red}},\inlinebox{\texttt{morning}} \\

    P&\ref{fig:qualitative::multi:::canoe}&\texttt{person} \inlinebox{\textcolor{white}{attr}} \texttt{at} \inlinebox{\textcolor{white}{attr}} \texttt{wearing an orange life vest}. & \inlinebox{\texttt{canoeing}},\inlinebox{\texttt{lake}} & \inlinebox{\texttt{kayaking}},\inlinebox{\texttt{river}}  \\

    \bottomrule

    \end{tabular}
    }
    \vspace{-1em}
\end{table*}

\noindent
\textbf{Concept bank attributes}. \cref{tab:prompts} presents learned vectors of concept/attribute modulations in context to the prompts used. In total, 33 unique modulations are presented in the paper, of which 10 relate to active objects, 11 are based on the action performed, 8 on scene changes, and 4 include object/action/scene combinations. With respect to the models used, 12 unique modulations are presented for Video LLaMA 3 logits, 11 for Gemma 3, and 10 for Phi 4 MM. The inclusion of different modulations shows the wide applicability of \serpantinelight{TRANSPORTER} and its usability as a tool for exploring different aspects of video understanding.

\section{OT types and compute times}
\label{sec:heuristic_ot}

\serpantinelight{TRANSPORTER} is based on an optimal transport coupling between embedding spaces. Dimension size differences between representations limit the potential application of standard optimal transport approaches. As an alternative to the proposed $\rho$-OT module in~\cref{sec:method::description}, other learned approaches such as Wasserstein Auto-Encoders (WAE)~\cite{tolstikhin2018wasserstein}, and Gromov-Wasserstein Auto-Encoders (GWAE)~\cite{nakagawagromov} can also be used. \cref{tab:ot_metrics} presents FVD video quality across optimal transport methods as well as their computational overheads. Performance drops significantly for the heuristic approach as the transport plan in sGW does not preserve the sequential structure of tokens and their dynamics. The autoencoder~\cite{tolstikhin2018wasserstein,nakagawagromov} approaches are implemented with two-MLP-layer encoders and decoders to avoid OOM errors. They achieve comparable performance to the proposed $\rho$-OT, but with a significant increase in computations required per iteration. Overall, $\rho$-OT with $P=100$ provides a balanced OT approach given the compute overhead and performance.

\section{Ablations with Phi 4 MM logits}
\label{sec:precision_recall}

Supplementary to~\cref{tab:transporter_ablations}, ablations are performed on \serpantinelight{TRANSPORTER} settings on Phi 4 MM logits.~\cref {tab:transporter_ablations_phi4} presents results over both coupling network and concept banks variants. Caption similarity drops in the inference-only setting compared to any of the other coupling network alternatives. Similarly, to~\cref{tab:transporter_ablations}, a small divergence is observed between the number of projections $P$. Overall,  $P=100$ still maintains competitive caption similarity with the average difference between the two top-performing projection settings being $\pm0.71$ across BLEU scores, $\pm0.91$ for CIDEr, $\pm0.48$ for METEOR, and $\pm0.73$ for SPICE. For the concept bank, most significant improvements are observed for CIDEr with $+2.31$ and B@2 with $+1.83$.  

\begin{table}[t]
    \caption{\textbf{Video quality, semantic alignment, and compute times across OT methods}. Times are averaged over 1K videos across 3 runs on a single NVIDIA L40s with Video LLaMA 3 embeddings as targets. It is assumed that tensors are on the GPU without CPU offloading. Best results are in \textbf{bold} and the used setting is in \textcolor{cvprblue}{blue}.}
    \centering
    \resizebox{\linewidth}{!}{%
    \setlength\tabcolsep{1.4pt}
    \begin{tabular}{cl c rrr}
        \toprule
         \multicolumn{2}{c}{\multirow{2}{*}{OT Method}} & \multirow{2}{*}{FVD$\downarrow$} & \multicolumn{3}{c}{Compute time (secs. [secs./it])$\downarrow$} \\
         && &\multicolumn{1}{c}{fwd} & \multicolumn{1}{c}{bwd} & \multicolumn{1}{c}{tot} \\
         \multicolumn{6}{l}{ \centerrule{3cm} Heuristic \centerrule{5.45cm}}\\
         sGW $P =$ & 100 & 5.42e$^{\!2}$ & 272.48 [2.18] & \multicolumn{1}{c}{N/A} &  272.48 [2.18]\\

         \multicolumn{6}{l}{ \centerrule{3cm} Learned \centerrule{5.6cm}}\\
         \multirow{2}{*}{WAE\textcolor{red}{$^\dagger$} $C= \Bigl\{$} & 768 & 1.73e$^{\!2}$ & 538.81 [4.31] & 178.67 [1.43] & 717.43 [5.74]\\
         & 1152 & 1.46e$^{\!2}$ & 778.84 [6.23] & 234.86 [1.88] & 1012.03 [8.11]\\
         GWAE\textcolor{red}{$^\dagger$} $C =$ & 1152 &  1.50e$^{\!2}$ & 853.73 [6.83] & 217.55 [1.74] & 1071.24 [8.57]\\
         \multicolumn{6}{l}{ \centerrule{3cm} \serpantinelight{TRANSPORTER} \centerrule{4.17cm}}\\
         \multirow{4}{*}{$\rho$-OT $P= \Biggl\{$} & 50 & 1.62e$^{\!2}$ & \textbf{206.23} [\textbf{1.65}] & \textbf{49.97} [\textbf{0.40}] & \textbf{256.23} [\textbf{2.05}]\\
         & \textcolor{cvprblue}{100} & \textcolor{cvprblue}{1.25e$^{\!2}$} & \textcolor{cvprblue}{227.64} [\textcolor{cvprblue}{1.82}] & \textcolor{cvprblue}{65.11} [\textcolor{cvprblue}{0.52}] & \textcolor{cvprblue}{292.46} [\textcolor{cvprblue}{2.34}]\\
         & 200 & 1.09e$^{\!2}$ & 260.03 [2.08] & 126.25 [1.01] & 386.24 [3.09]\\
         & 400 & \textbf{9.81e}$^{\!\mathbf{1}}$ & 361.27 [2.89] & 157.52 [1.26] & 518.76 [4.15]\\
         \bottomrule
    \end{tabular}
    }
    \label{tab:ot_metrics}
    \vspace{-1em}
\end{table}

\section{Modulations across phenomena}
\label{sec:linguistic_qualitative}

Critically, the applicability of VLMs depends on their ability to distinguish between opposing or distinct semantic meanings. Prior benchmarks have explored VLMs' response sensitivity across object plurality~\cite{rahmanzadehgervi2024vision}, preposition~\cite{parcalabescu2022valse}, and plausibility~\cite{kamath2023s}. In parallel with these studies on VLM outputs, \serpantinelight{TRANSPORTER} can provide visual representations of token predictions. Such modulation pairs linguistically ground VLMs across various aspects. This provides a novel path for future work to visually explore interpretability-related themes.

\begin{figure*}[t]
    \begin{subfigure}[t]{\textwidth}
    \vspace{-1em}
    \includegraphics[width=\textwidth,clip]{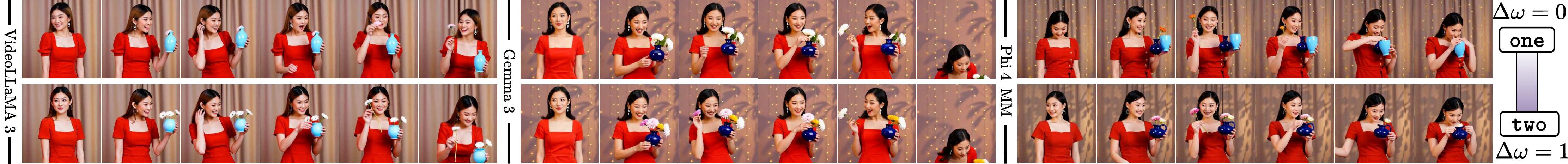}
    \vspace{-1em}
    \caption{\textbf{plurality/counting}: \texttt{A woman with a red dress looks at a blue vase and picks} \inlinebox{\texttt{one}}|\inlinebox{\texttt{two}} \texttt{flowers out of \underline{four}}.}
    \label{fig:modulation_types::cross_model:a}
    \end{subfigure}
    \begin{subfigure}[t]{\textwidth}
    \includegraphics[width=\textwidth,clip]{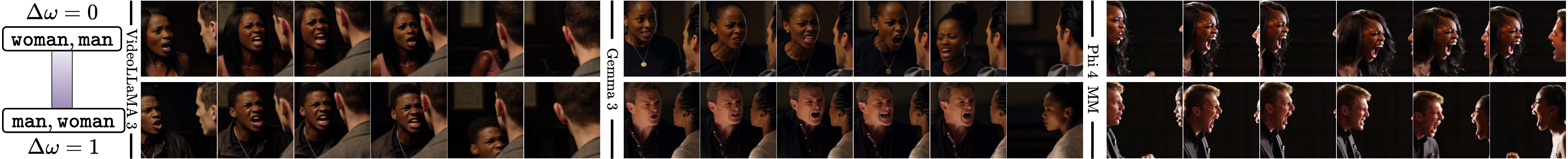}
    \vspace{-1em}
    \caption{\textbf{preposition}: \texttt{A close-up scene of a} \inlinebox{\texttt{woman}}|\inlinebox{\texttt{man}} \texttt{shouting intensively to a} \inlinebox{\texttt{man}}|\inlinebox{\texttt{woman}} \texttt{as \underline{she} walks way}.}
    \label{fig:modulation_types::cross_model:b}
    \end{subfigure}
    \begin{subfigure}[t]{\textwidth}
    \includegraphics[width=\textwidth,clip]{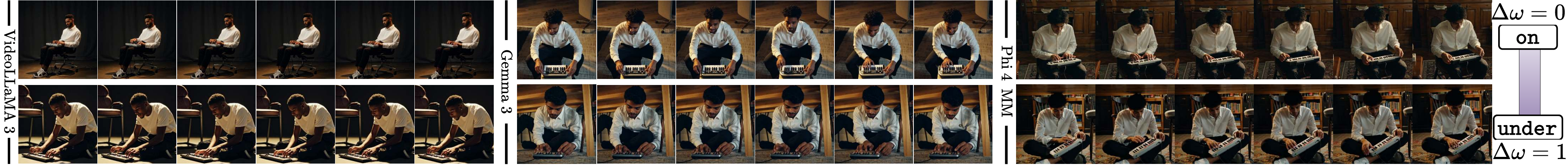}
    \vspace{-1em}
    \caption{\textbf{relations/plausability}: \texttt{A well-dressed man sitting} \inlinebox{\texttt{on}}|\inlinebox{\texttt{under}} \texttt{a \underline{chair} playing with a small electronic keyboard}.}
    \label{fig:modulation_types::cross_model:c}
    \end{subfigure}
    \captionof{figure}{\textbf{Modulation types comparisons} across VideoLLaMA 3, Gemma 3, Phi 4 MM, over (a) \textbf{plurarity/counting}, (b) \textbf{preposition}, and (c) \textbf{relations/plausibility}. The same rng (0) is used for \serpantinelight{TRANSPORTER} inference across VLMs.}
\label{fig:rebuttal:cross_model}
\end{figure*}

\noindent
\textbf{Plurality/counting}.~\cref{fig:modulation_types::cross_model:a} presents modulation over \inlinebox{\texttt{one}} and \inlinebox{\texttt{two}} flowers being picked.
VLMs often struggle with counting since it requires semantically
matching language to visual prompts, both at a semantic level and in terms of location in frames. Videos present an additional challenge, as transitioning from \inlinebox{\texttt{one}} to \inlinebox{\texttt{two}} flowers requires an extra action that not all models can always infer.

\noindent
\textbf{Preposition}.~\cref{fig:modulation_types::cross_model:b} visualizes a linguistic preposition, which should result in significant changes in the scene. The initial \inlinebox{\texttt{woman}} \texttt{shouting intensively to a} \inlinebox{\texttt{man}} is not successfully changed to \inlinebox{\texttt{man}} \texttt{shouting intensively to a} \inlinebox{\texttt{woman}}. The resulting videos from VideoLLaMA 3 and Phi 4 MM show that such fine-grained preposition details are often missed.

\noindent
\textbf{Relation/plausibility}.~\cref{fig:modulation_types::cross_model:c} shows VLMs' reliance on priors in terms of object relations and physical plasibilities. Resulting videos when modulating to \texttt{sitting} \inlinebox{\texttt{under}} \texttt{a chair} either visualize the chair at a different location in the scene (lack of spatial understanding) or the scene resembles \texttt{sitting} \inlinebox{\texttt{under}} \texttt{a table} (potentially closer to the training distribution).

\section{Additional qualitative results}
\label{sec:qualitative}

In addition to the examples in~\cref{sec:experiments::examples}, modulations for objects, actions, scenes, and their combinations are shown.

\noindent
\textbf{Active object}. As in~\cref{fig:qualitative::object:::pink,fig:qualitative::object:::wood}, \serpantinelight{TRANSPORTER} can generate videos to visualize transitions between different object attributes, such as color or material types. The sharpness of the transition also varies across modulations. In fine-grained object changes such as \inlinebox{\texttt{starfish}} to \inlinebox{\texttt{seashell}} in~\cref{fig:qualitative::object:::starfish} or \inlinebox{\texttt{roll}} to \inlinebox{\texttt{brush}} in~\cref{fig:qualitative::object:::roll}, the transition from $\pi^-$ to $\pi^+$ happens over small $\Delta \omega$ fractions. Learned geometric and appearance object properties can also be seen during transitions in ~\cref{fig:qualitative::object:::cookies,fig:qualitative::object:::book} with edges in complex shapes such as those of \inlinebox{\texttt{christmas tree}} being flattened to round edges, or the hardback of a \inlinebox{\texttt{book}} transitioning to a paperback, and eventually \inlinebox{\texttt{newspaper}} pages.
\blfootnote{\textcolor{red}{$^\dagger$} Trained on half batch size and double gradient accumulation steps due to OOM. Approx. fwd/bwd/tot reported.}

\noindent
\textbf{Action performance}. In actions where their performance is proximal, such as \inlinebox{\texttt{surfing}} and \inlinebox{\texttt{kitesurfing}} in~\cref{fig:qualitative::action:::surf}, \inlinebox{\texttt{roller skating}} and \inlinebox{\texttt{skateboarding}} in~\cref{fig:qualitative::action:::skate}, and \inlinebox{\texttt{toe bouncing}} and \inlinebox{\texttt{head bouncing}} in~\cref{fig:qualitative::action:::bounce}, modulations show that small and distinct changes are learned. The difference in speed can be seen in cases where action verbs differ significantly in their executions, such as \inlinebox{\texttt{running}} and \inlinebox{\texttt{spinning}} in~\cref{fig:qualitative::action:::run}. Transitions between actions with even larger differences in their executions are simultaneously visible as in \inlinebox{\texttt{riding}} to \inlinebox{\texttt{leading}} in~\cref{fig:qualitative::action:::ride}.

\begin{table}[t]
    \caption{\textbf{Phi 4 MM cosine similarity} ($\nabla\!\text{cos}$), \textbf{BLEU (B@1, B@2, B@3, B@4), CIDEr (C), METEOR (M), and SPICE (S)} over captions from \serpantinelight{TRANSPORTER} videos and target captions. Settings are grouped in relation to the coupling network or the concept bank. Best results are \textbf{bold} and second best are \underline{underlined}.}
    \centering
    \resizebox{\linewidth}{!}{%
    \setlength\tabcolsep{1.8pt}
    \begin{tabular}{ll cccccccc}
        \toprule
         \multicolumn{2}{l}{\multirow{2}{*}{Method}} & \multicolumn{8}{c}{Phi 4 MM} \\ 
         \cmidrule(lr){3-10} 
         && $\nabla\!\cos$  & B@1 & B@2 & B@3 & B@4 & C & M & S \\
        \midrule
        \multicolumn{2}{l}{Baseline~\cite{stergiou2023leaping}} & 0.04 & 25.22 & 11.10 & 4.35 & 0.01 & 5.01 & 11.84 & 8.64 \\
        \addlinespace[0.1em]
        \multicolumn{2}{l}{\centerrule{3cm}} &\multicolumn{8}{l}{Coupling network ablations \centerrule{3.4cm}} \bstrut \\
        \multicolumn{2}{l}{inference-only} & 0.34 & 29.25 & 20.52 & 14.72 & 10.28 & 23.89 & 15.92 & 21.50 \\
        \multicolumn{2}{l}{$\text{mean}\langle\Phi_{\Omega_1,\Omega_2}\rangle$} & 0.37 & 30.93 & 21.23 & 15.45 & 11.64 & 24.11 & 16.63 & 24.37 \\
        \multicolumn{2}{l}{sGW OT}& 0.32 & 25.61 & 16.47 & 13.28 & 9.81 & 23.44 & 13.15 & 12.63 \\
        \multirow{2}{*}{$P=\Bigl\{$} & $50$ & 0.41 & 34.03 & 22.59 & 16.66 & 12.82 & 28.64 & 20.30 & 27.75 \\
         & $400$ & \textbf{0.44} & \textbf{35.24} & \textbf{23.78} & \textbf{18.53} & \textbf{14.08} & \textbf{30.42} & \textbf{21.35} & \textbf{28.96} \\
        \addlinespace[0.1em]
        \multicolumn{2}{l}{\centerrule{3cm}} &\multicolumn{8}{l}{Concept bank ablation \centerrule{4.1cm}} \bstrut \\
        $\Delta\omega$ & JS & 0.36 & 33.15 & 21.42 & 15.87 & 11.93 & 28.20 & 18.56 & 27.42 \\
        \midrule
        \multicolumn{2}{l}{\serpantinelight{TRANSPORTER}} & \underline{0.43} & \underline{34.77} & \underline{23.25} & \underline{17.30} & \underline{13.49} & \underline{29.51} & \underline{20.87} & \underline{28.23} \\
        \bottomrule
    \end{tabular}
    }
    \label{tab:transporter_ablations_phi4}
    \vspace{-1em}
\end{table}

\noindent
\textbf{Scene context}. Scene modulations include backgrounds~\cref{fig:qualitative::scene:::parking,fig:qualitative::scene:::outdoor,fig:qualitative::scene:::tree,fig:qualitative::scene:::sky}, and locations~\cref{fig:qualitative::scene:::sandtrap,fig:qualitative::scene:::clay}.  The resulting videos show \serpantinelight{TRANSPORTER}'s ability to generate videos with in-context modulations. Aspects such as the performance of the action or the appearances of objects/actors remain constant throughout the transition between source and target logits.

\noindent
\textbf{Combined}. In settings with combined modulation, transitions of individual logit pairs are visualized over different $\Delta \omega$. \cref{fig:qualitative::multi:::bike} shows that transitioning between \inlinebox{\texttt{black}} and \inlinebox{\texttt{blue}} happens first, as their logit and embedding distance is shorter than other pairs. In contrast, the changes from \inlinebox{\texttt{jacket}} to \inlinebox{\texttt{t-shirt}}, and \inlinebox{\texttt{bicycle}} to \inlinebox{\texttt{scooter}} happen at similar $\Delta \omega$ variance, showing that appearance attributes are encoded over larger manifolds.

\section{Further discussions}
\label{sec:discussion}

\noindent
\textbf{Limitations}. The manually selected modulations across model logits aim to visualize the learned representations and manifold between concepts. However, as noted, the interpolation is done based on learned vectors for concept pairs over VLM logit divergences. This can limit the clarity of explanations of entangled concepts for which vectors $\mathbf{q}_o$ are learned from source and target concept/attributes that are not directly orthogonal. As the current method is based on discrete concept pairs, a future direction could be generalizing L2V and \serpantinelight{TRANSPORTER} to continuous attributes. For example, instead of \inlinebox{\texttt{slow}} $\leftrightarrow$ \inlinebox{\texttt{fast}}, the model could learn a \inlinebox{\texttt{speed}} vector, allowing a user to control the pace with a continuous scalar, not just an interpolation. This can further enable the applicability of \serpantinelight{TRANSPORTER} in zero-shot concept pairs at inference.  

\noindent
\textbf{Potential improvements}. Although the focus has been on establishing L2V as a fruitful direction for video model interpretability, extending the method to few-shot video concept visualization can be an area of improvement. Instead of a discrete, static concept bank $\mathbf{Q}$, L2V can be formulated to model concept directions from a (learned) continuous function $f_{\theta}(\pi^{-}, \pi^{+}) \rightarrow \mathbf{q}_o$. This function could be trained to approximate the embedding manifold of any two text prompts used as input and output the corresponding latent direction vector $\mathbf{q}$. Potentially, this could be modeled contrastively~\cite{tschannen2025siglip,caron2020unsupervised} on a large corpus of text-pair-video data. Inference-time semantic edits~\cite{dalva2025fluxspace}, uncertainty quantification~\cite{ye2024benchmarking}, and model merging with model-specific vectors~\cite{goddard2024arcee,wang2024sam} can also be adopted to reduce reliance on training concept vectors.

\begin{figure*}[ht]
    \label{fig:qualitative::object}
    \centering
    \begin{subfigure}[t]{\textwidth}
        \centering
        \includegraphics[width=\linewidth]{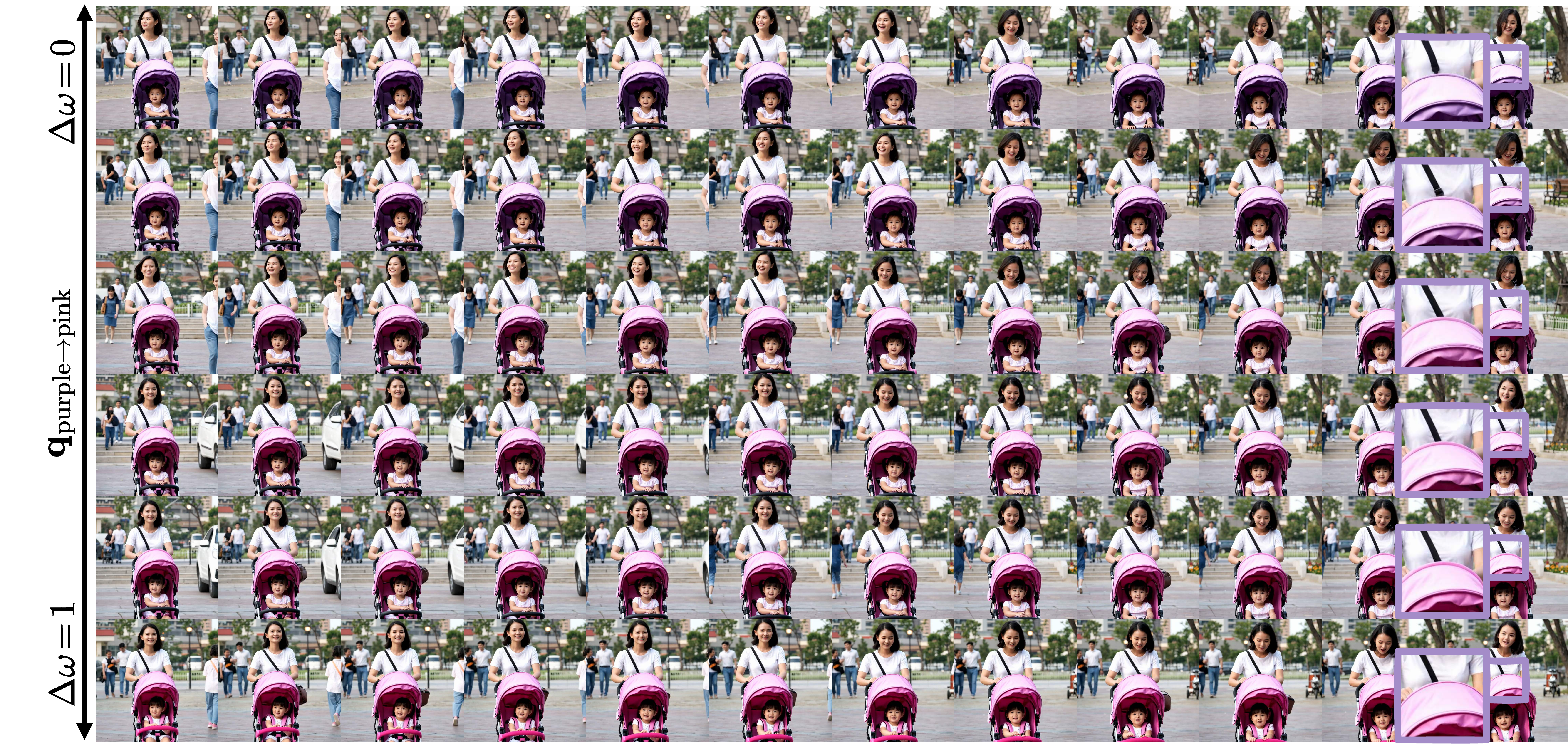}
        \caption{Video LLaMA 3 modulations between \inlinebox{\texttt{purple}} and \inlinebox{\texttt{pink}}.}
        \label{fig:qualitative::object:::pink}
    \end{subfigure} 
    \begin{subfigure}[t]{\textwidth}
        \centering
        \includegraphics[width=\linewidth]{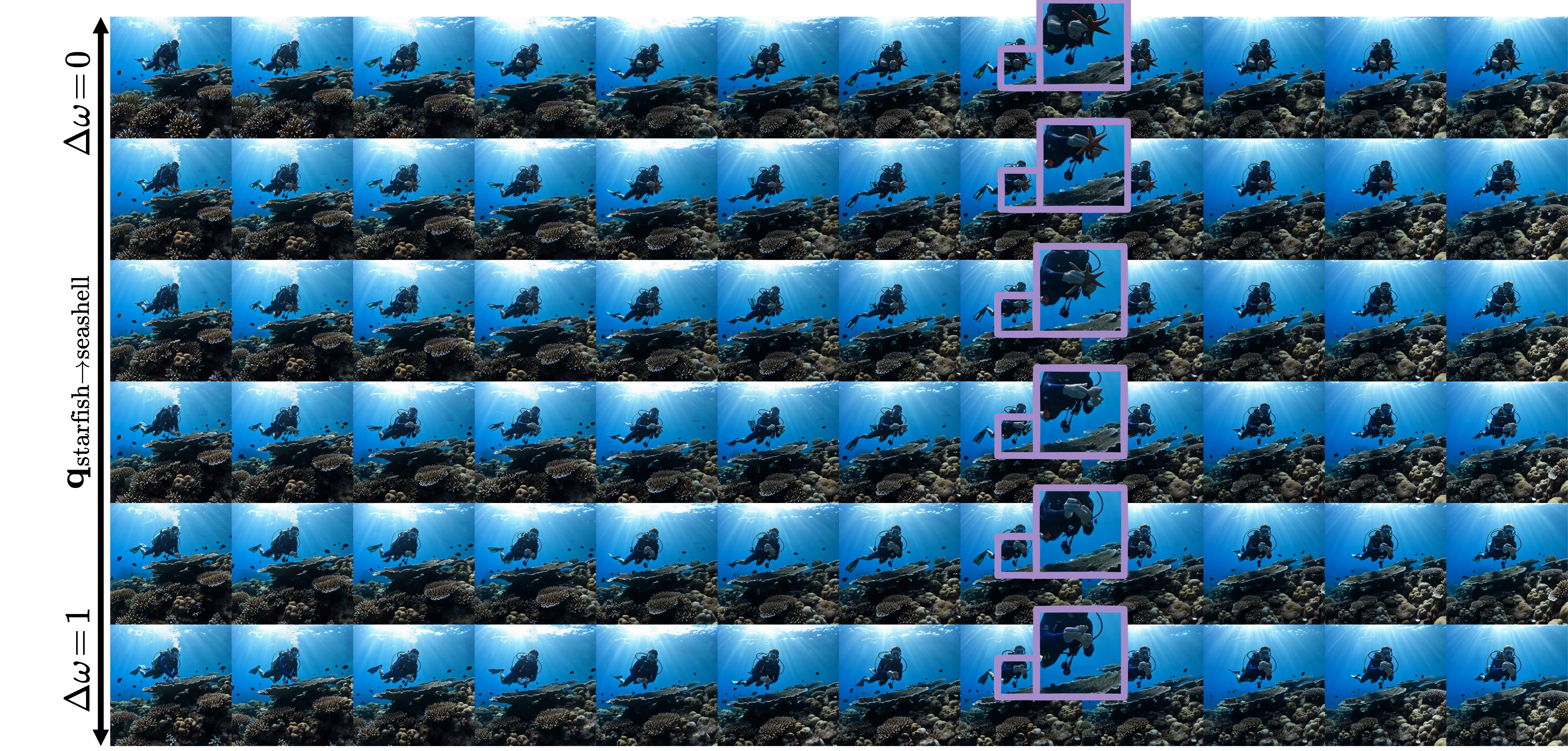}
        \caption{Video LLaMA 3 modulations between \inlinebox{\texttt{starfish}} and \inlinebox{\texttt{seashell}}.}
        \label{fig:qualitative::object:::starfish}
    \end{subfigure}
    \caption{\textbf{Examples of active object modulations}. Frame quality is compressed due to filesize (best viewed digitally).}
\end{figure*}
\begin{figure*}[ht]\ContinuedFloat
    \centering
    \begin{subfigure}[t]{\textwidth}
        \centering
        \includegraphics[width=\linewidth]{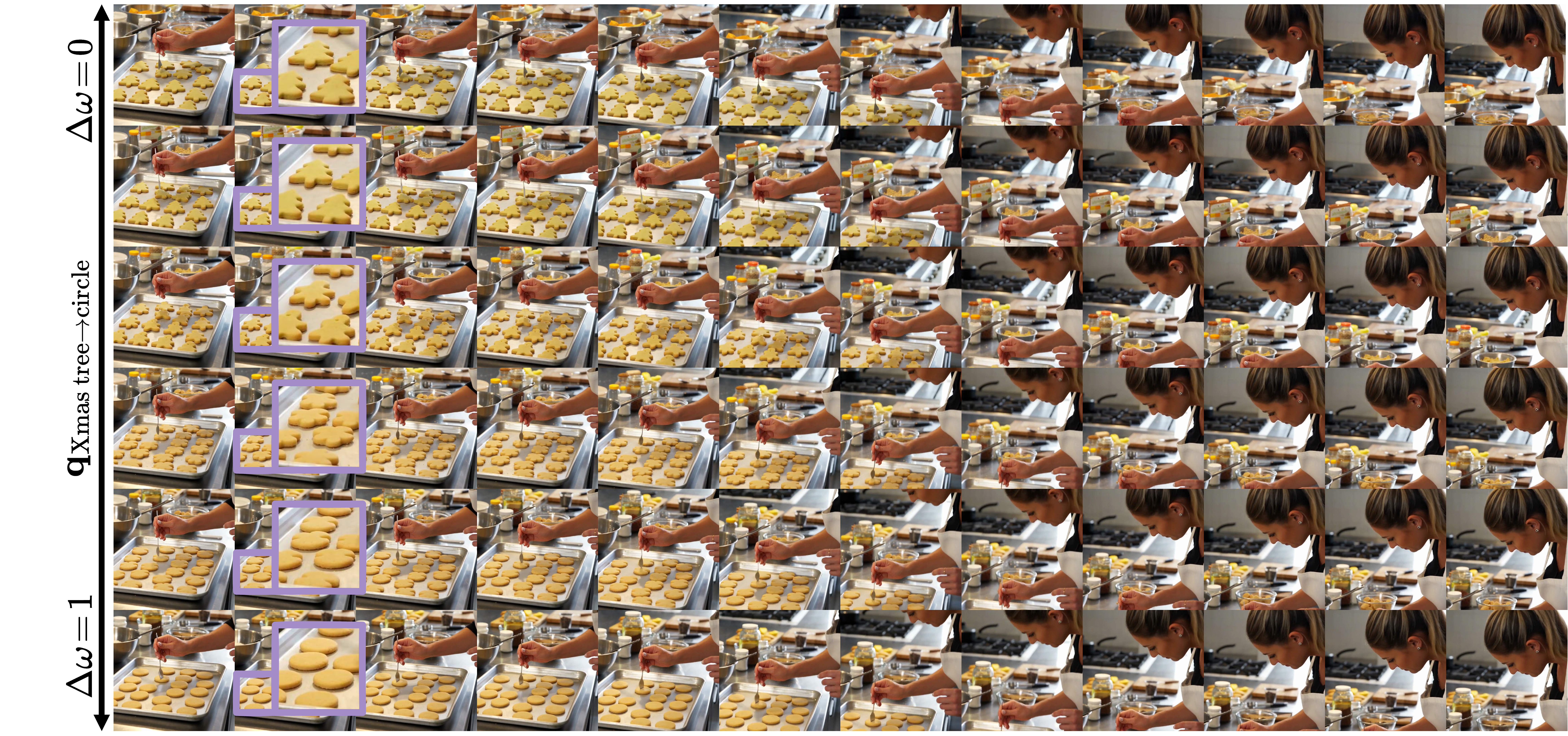}
        \caption{Video LLaMA 3 modulations between \inlinebox{\texttt{christmas tree}} and \inlinebox{\texttt{circle}}.}
        \label{fig:qualitative::object:::cookies}
    \end{subfigure}
    \begin{subfigure}[t]{\textwidth}
        \centering
        \includegraphics[width=\linewidth]{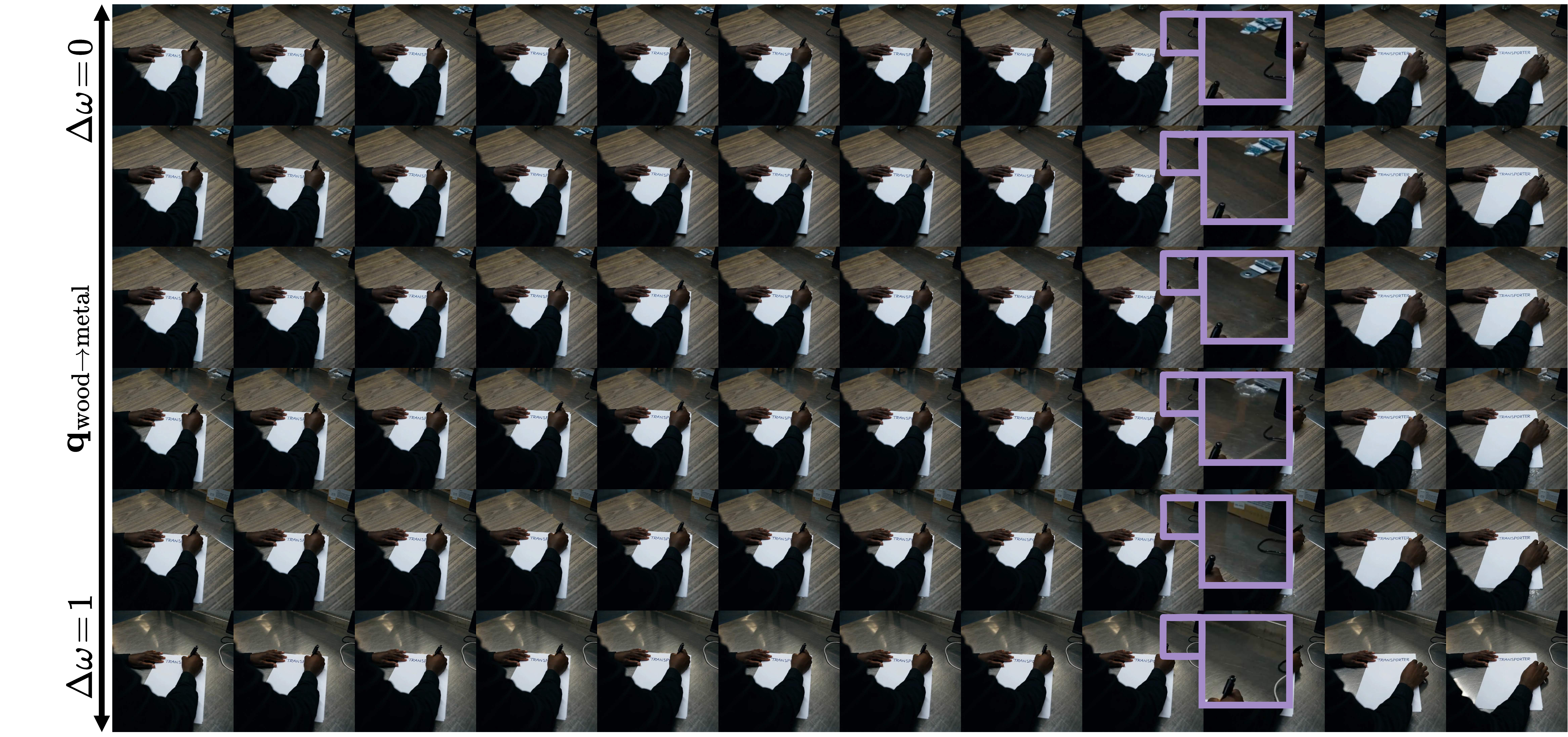}
        \caption{Gemma 3 modulations between \inlinebox{\texttt{wood}} and \inlinebox{\texttt{metal}}.}
        \label{fig:qualitative::object:::wood}
    \end{subfigure}
    \caption[]{\textbf{Examples of active object modulations}. Frame quality is compressed due to filesize (best viewed digitally).}
\end{figure*}
\begin{figure*}[ht]\ContinuedFloat
    \centering
    \begin{subfigure}[t]{\textwidth}
        \centering
        \includegraphics[width=\linewidth]{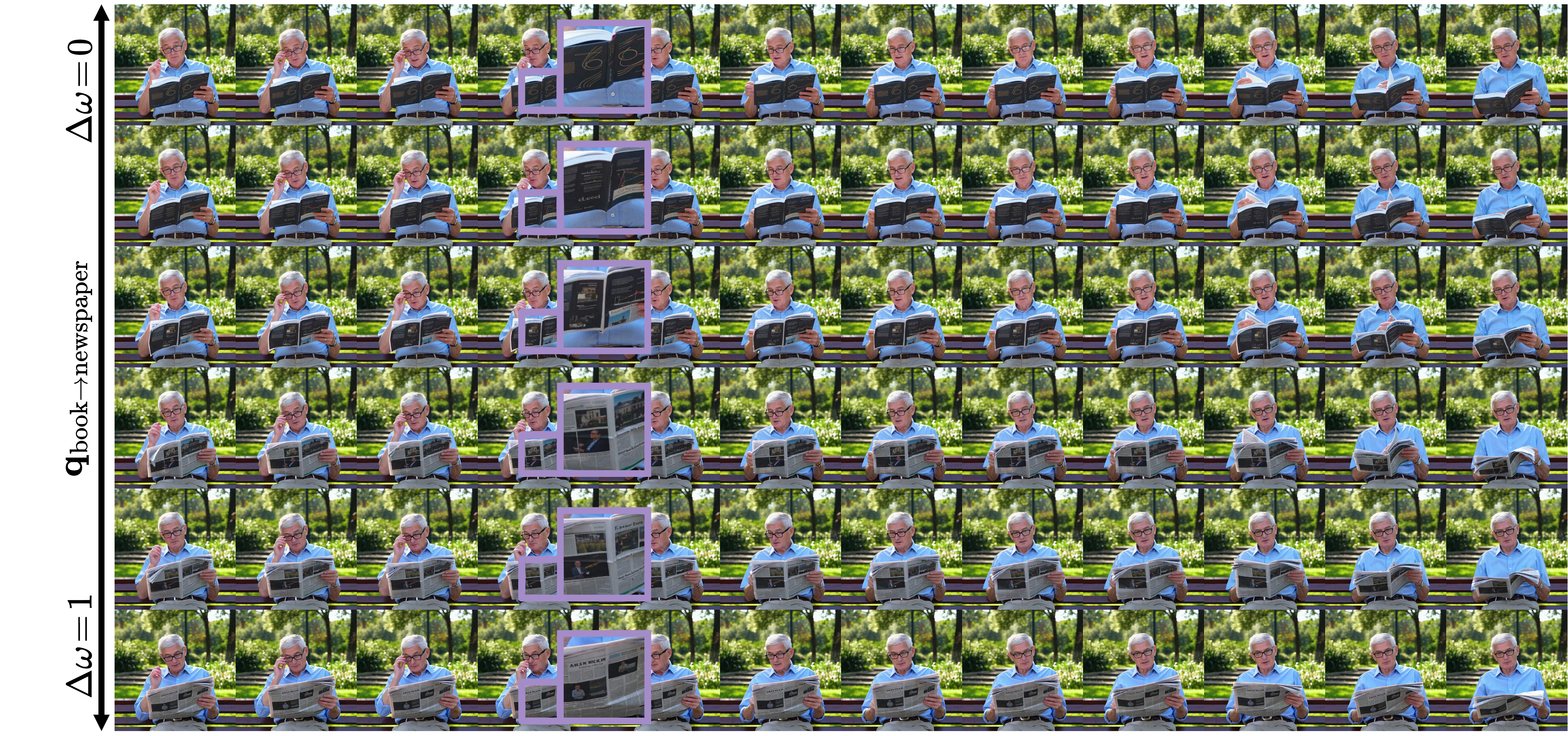}
        \caption{Gemma 3 modulations between \inlinebox{\texttt{book}} and \inlinebox{\texttt{newspaper}}.}
        \label{fig:qualitative::object:::book}
    \end{subfigure}
    \begin{subfigure}[t]{\textwidth}
        \centering
        \includegraphics[width=\linewidth]{figs/sup_mat/qualitative_objects/TRANSPORTER_v1_qualitative_objects-mango_apple.pdf}
        \caption{Gemma 3 modulations between \inlinebox{\texttt{mango}} and \inlinebox{\texttt{apple}}.}
        \label{fig:qualitative::object:::mango}
    \end{subfigure}
    \caption[]{\textbf{Examples of active object modulations}. Frame quality is compressed due to filesize (best viewed digitally).}
\end{figure*}
\begin{figure*}[ht]\ContinuedFloat
    \centering
    \begin{subfigure}[t]{\textwidth}
        \centering
        \includegraphics[width=\linewidth]{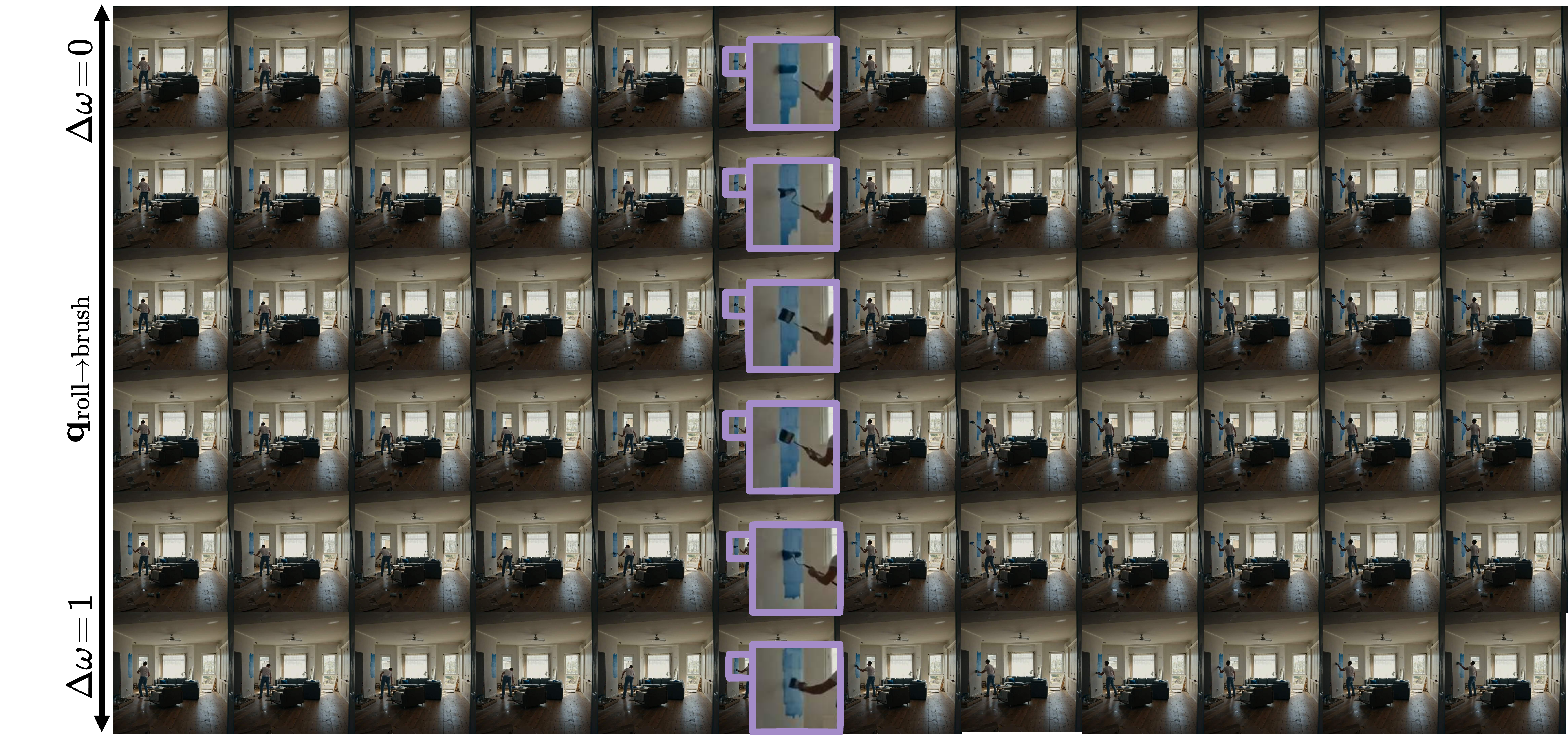}
        \caption{Phi 4 MM modulations between \inlinebox{\texttt{roll}} and \inlinebox{\texttt{brush}}.}
        \label{fig:qualitative::object:::roll}
    \end{subfigure}
    \begin{subfigure}[t]{\textwidth}
        \centering
        \includegraphics[width=\linewidth]{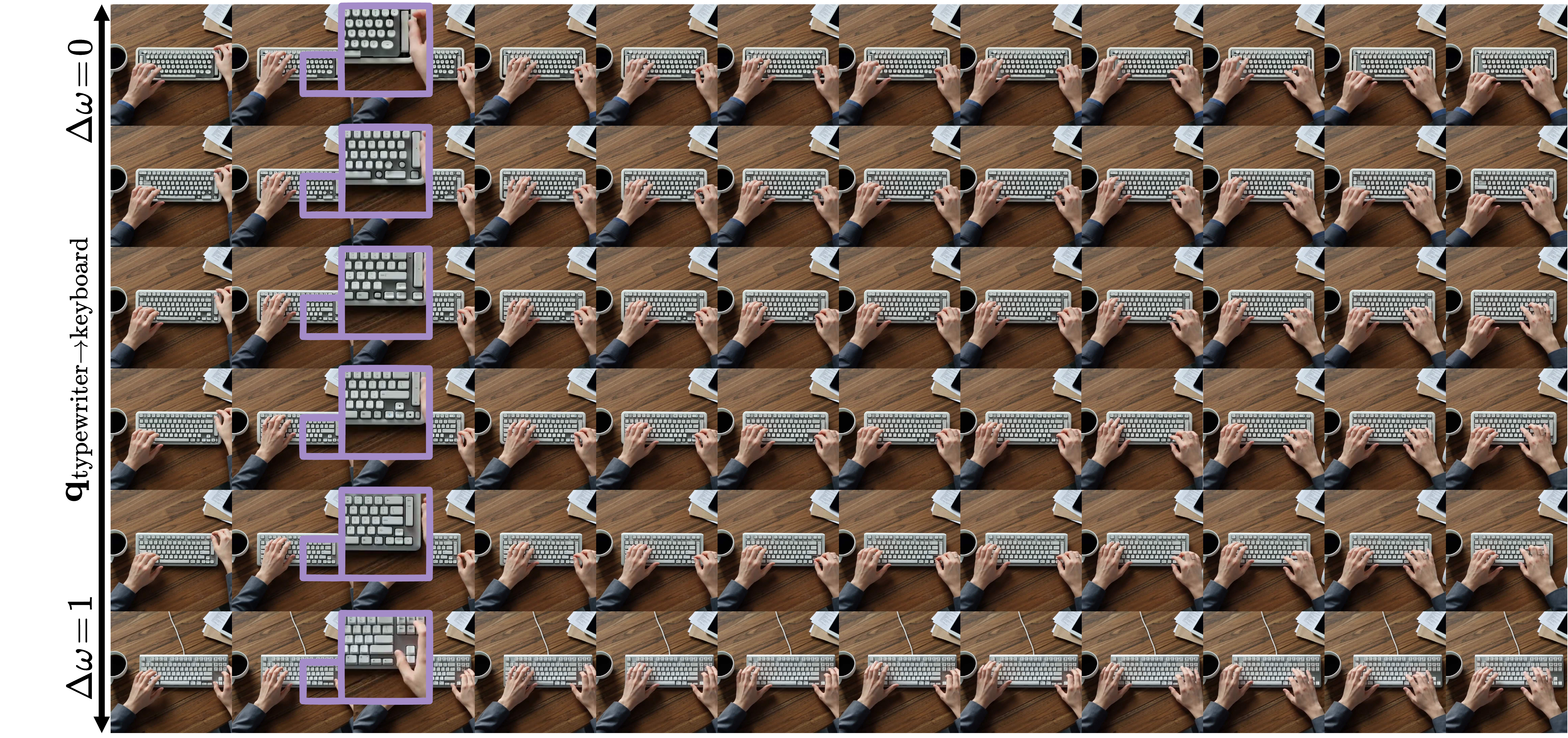}
        \caption{Phi 4 MM modulations between \inlinebox{\texttt{typewriter}} and \inlinebox{\texttt{keyboard}}.}
        \label{fig:qualitative::object:::typewriter}
    \end{subfigure}
    \caption[]{\textbf{Examples of active object modulations}. Frame quality is compressed due to filesize (best viewed digitally).}
\end{figure*}

\begin{figure*}[ht]
    \centering
    \label{fig:qualitative::action}
    \begin{subfigure}[t]{\textwidth}
        \centering
        \includegraphics[width=\linewidth]{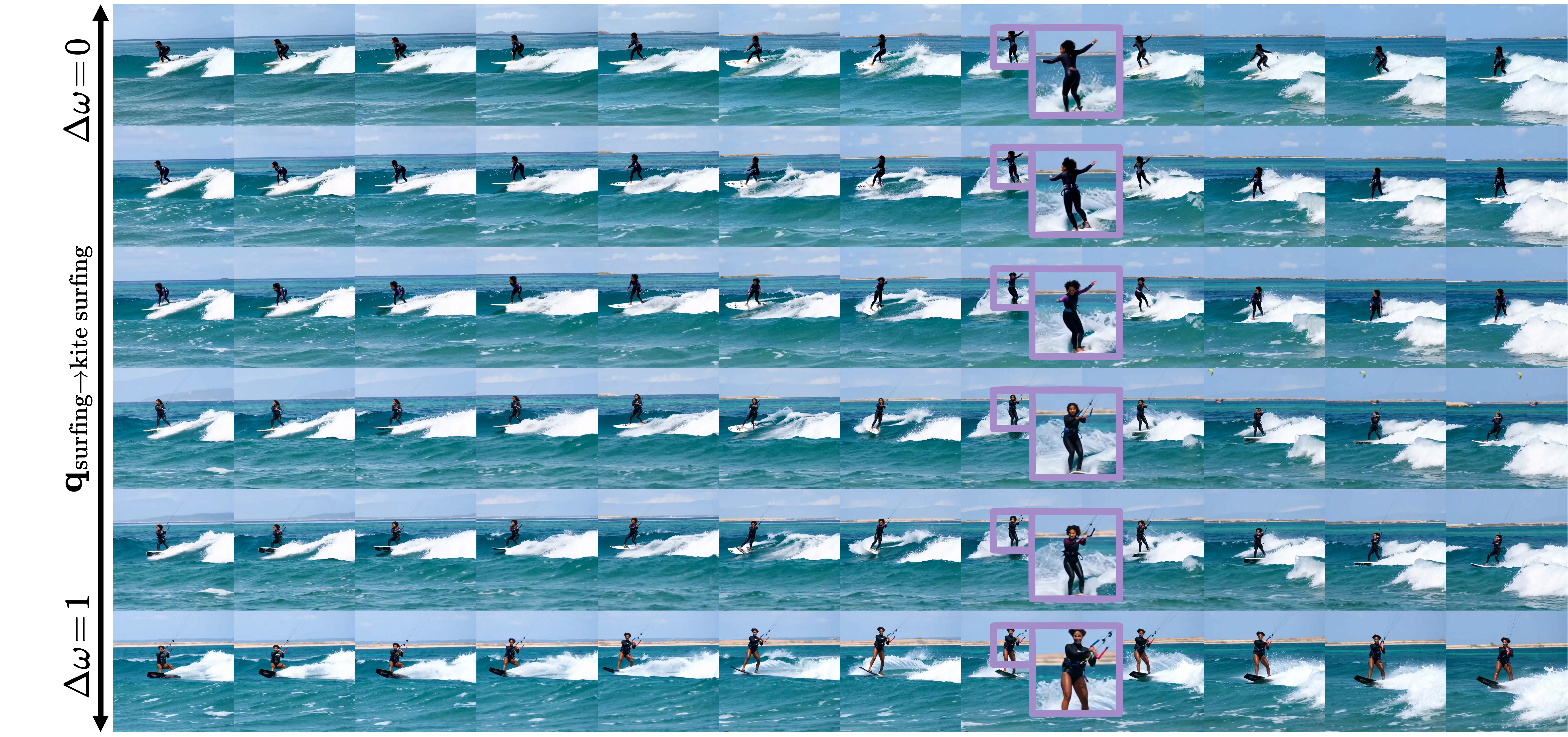}
        \caption{Video LLaMA 3 modulations between \inlinebox{\texttt{surfing}} and \inlinebox{\texttt{kitesurfing}}.}
        \label{fig:qualitative::action:::surf}
    \end{subfigure} 
    \begin{subfigure}[t]{\textwidth}
        \centering
        \includegraphics[width=\linewidth]{figs/sup_mat/qualitative_action/TRANSPORTER_v1_qualitative_action-roller_skating_skateboarding}
        \caption{Video LLaMA 3 modulations between \inlinebox{\texttt{roller skating}} and \inlinebox{\texttt{skateboarding}}.}
        \label{fig:qualitative::action:::skate}
    \end{subfigure}
    \caption{\textbf{Examples of action modulations}. Frame quality is compressed due to filesize (best viewed digitally).}
\end{figure*}
\begin{figure*}[ht]\ContinuedFloat
    \centering
    \begin{subfigure}[t]{\textwidth}
        \centering
        \includegraphics[width=\linewidth]{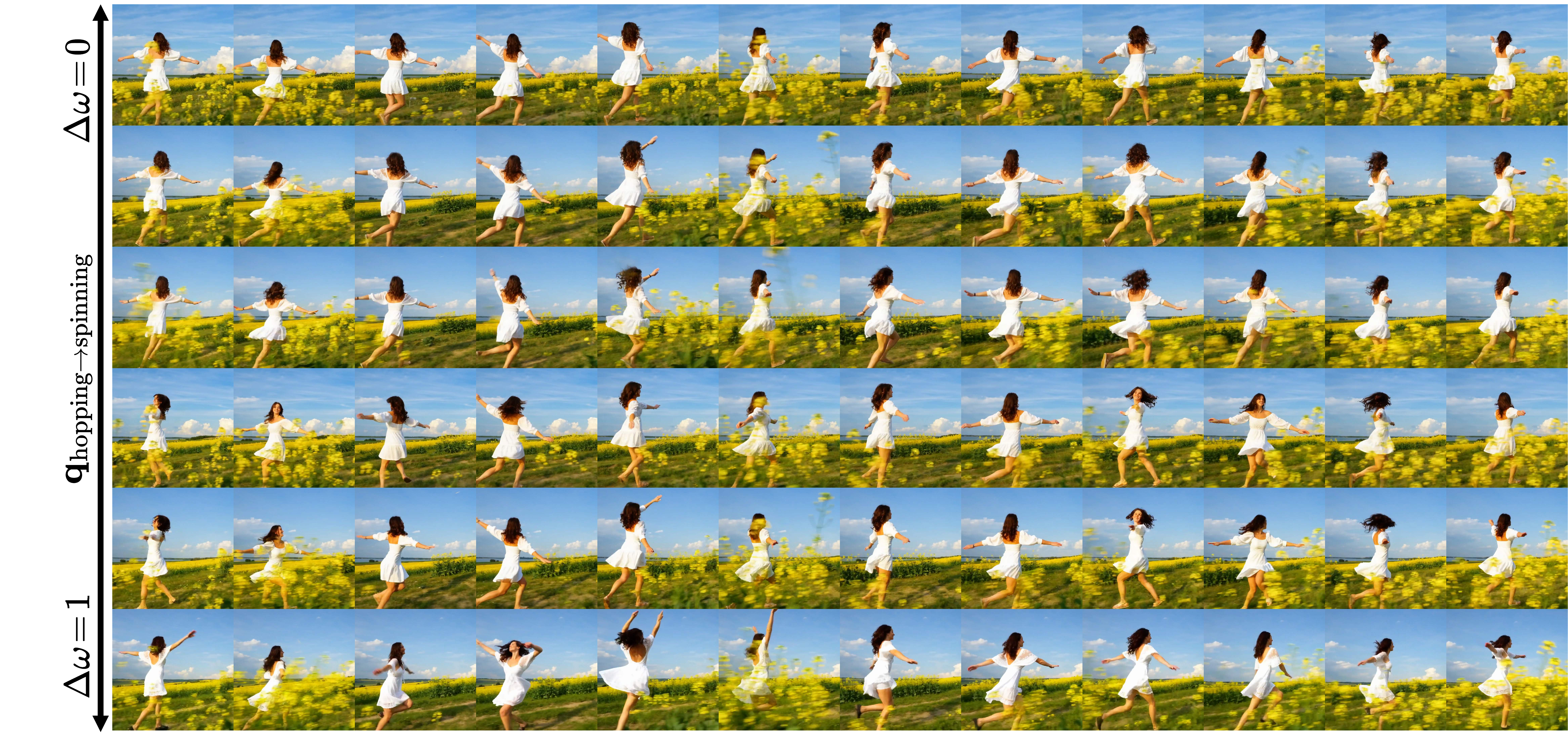}
        \caption{Video LLaMA 3 modulations between \inlinebox{\texttt{running}} and \inlinebox{\texttt{spinning}}.}
        \label{fig:qualitative::action:::run}
    \end{subfigure}
    \begin{subfigure}[t]{\textwidth}
        \centering
        \includegraphics[width=\linewidth]{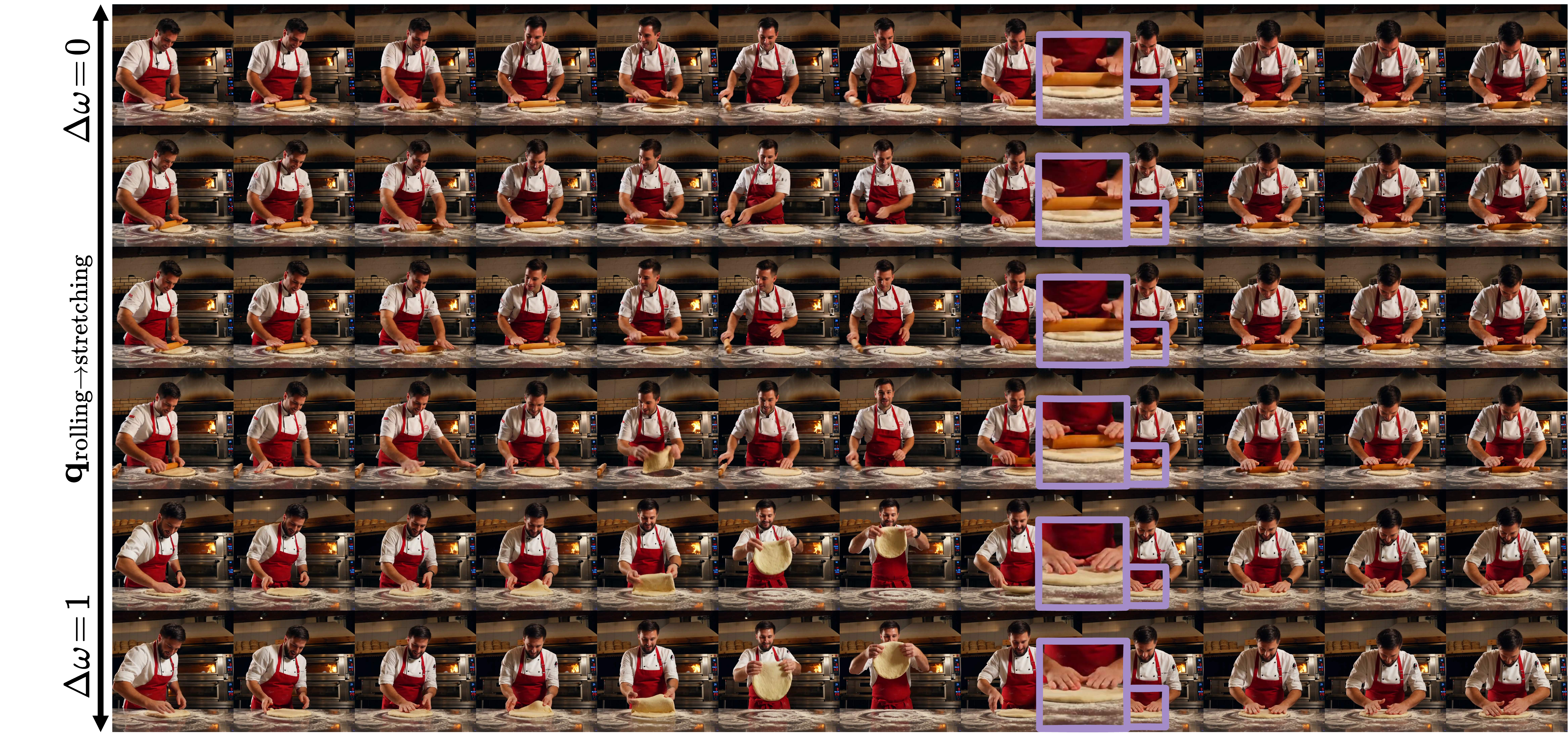}
        \caption{Gemma 3 modulations between \inlinebox{\texttt{rolling}} and \inlinebox{\texttt{stretching}}.}
        \label{fig:qualitative::action:::roll} 
    \end{subfigure}
    \caption[]{\textbf{Examples of action modulations}. Frame quality is compressed due to filesize (best viewed digitally).}
\end{figure*}
\begin{figure*}[ht]\ContinuedFloat
    \centering
    \begin{subfigure}[t]{\textwidth}
        \centering
        \includegraphics[width=\linewidth]{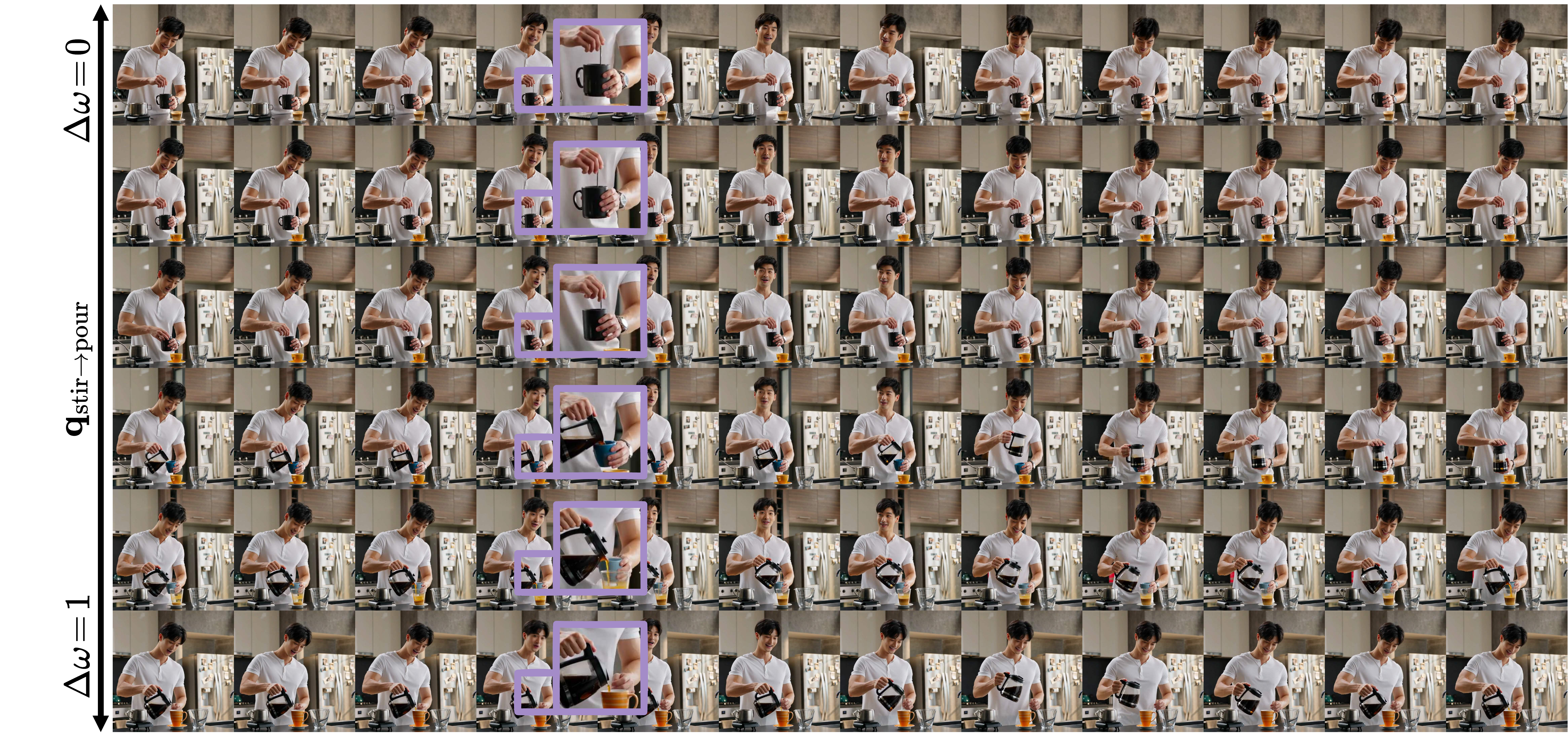}
        \caption{Gemma 3 modulations between \inlinebox{\texttt{stir}} and \inlinebox{\texttt{pour}}.}
        \label{fig:qualitative::action:::stir}
    \end{subfigure}
    \begin{subfigure}[t]{\textwidth}
        \centering
        \includegraphics[width=\linewidth]{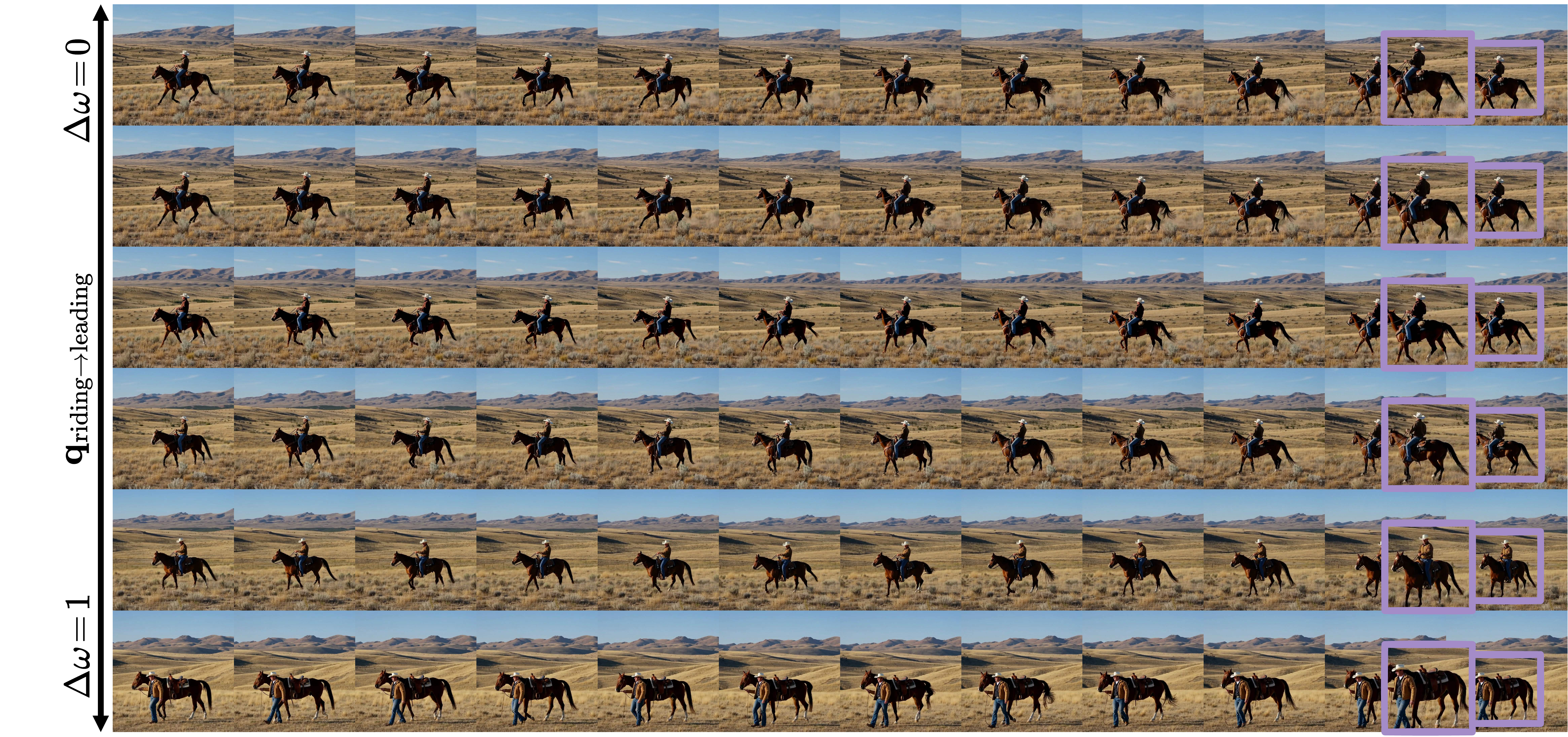}
        \caption{Gemma 3 modulations between \inlinebox{\texttt{riding}} and \inlinebox{\texttt{leading}}.}
        \label{fig:qualitative::action:::ride}
    \end{subfigure}
    \caption[]{\textbf{Examples of action modulations}. Frame quality is compressed due to filesize (best viewed digitally).}
\end{figure*}
\begin{figure*}[ht]\ContinuedFloat
    \centering
    \begin{subfigure}[t]{\textwidth}
        \centering
        \includegraphics[width=\linewidth]{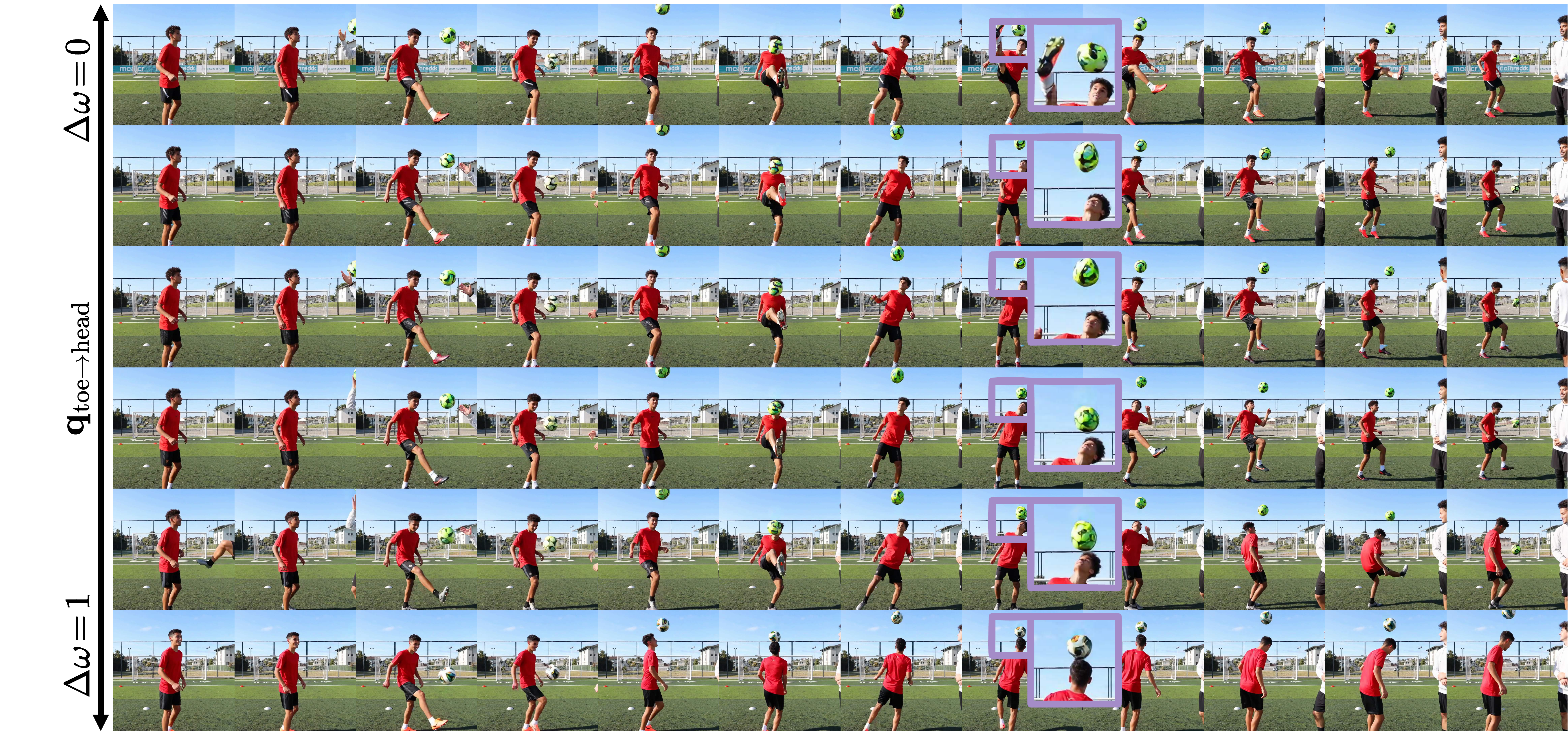}
        \caption{Phi 4 MM modulations between \inlinebox{\texttt{toe bouncing}} and \inlinebox{\texttt{head bouncing}}.}
        \label{fig:qualitative::action:::bounce}
    \end{subfigure}
    \begin{subfigure}[t]{\textwidth}
        \centering
        \includegraphics[width=\linewidth]{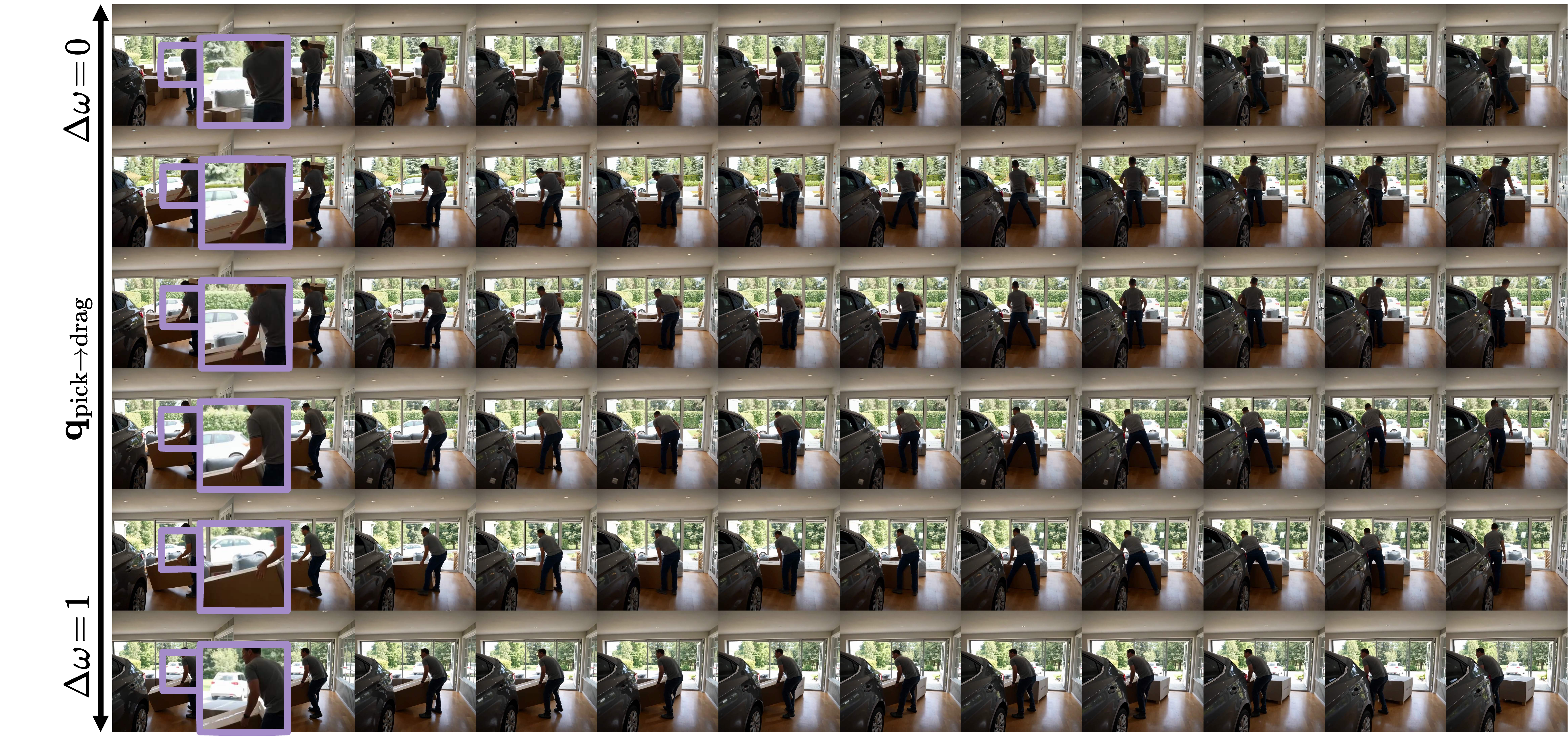}
        \caption{Phi 4 MM modulations between \inlinebox{\texttt{picks up}} and \inlinebox{\texttt{drags}}.}
        \label{fig:qualitative::action:::pick}
    \end{subfigure}
    \caption[]{\textbf{Examples of action modulations}. Frame quality is compressed due to filesize (best viewed digitally).}
\end{figure*}

\begin{figure*}[ht]
    \centering
    \label{fig:qualitative::scene}
    \begin{subfigure}[t]{\textwidth}
        \centering
        \includegraphics[width=\linewidth]{figs/sup_mat/qualitative_scene/TRANSPORTER_v1_qualitative_scene-parkinglot_underpass.pdf}
        \caption{Video LLaMA 3 modulations between \inlinebox{\texttt{parking lot}} and \inlinebox{\texttt{underpass}}.}
        \label{fig:qualitative::scene:::parking}
    \end{subfigure} 
    \begin{subfigure}[t]{\textwidth}
        \centering
        \includegraphics[width=\linewidth]{figs/sup_mat/qualitative_scene/TRANSPORTER_v1_qualitative_scene-grass_sandtrap.pdf}
        \caption{Video LLaMA 3 modulations between \inlinebox{\texttt{grass}} and \inlinebox{\texttt{sandtrap}}.}
        \label{fig:qualitative::scene:::sandtrap}
    \end{subfigure}
    \caption{\textbf{Examples of scene modulations}. Frame quality is compressed due to filesize (best viewed digitally).}
\end{figure*}
\begin{figure*}[ht]\ContinuedFloat
    \centering
    \begin{subfigure}[t]{\textwidth}
        \centering
        \includegraphics[width=\linewidth]{figs/sup_mat/qualitative_scene/TRANSPORTER_v1_qualitative_scene-outdoor_indoor.pdf}
        \caption{Gemma 3 modulations between \inlinebox{\texttt{outdoor}} and \inlinebox{\texttt{indoor}}.}
        \label{fig:qualitative::scene:::outdoor}
    \end{subfigure}
    \begin{subfigure}[t]{\textwidth}
        \centering
        \includegraphics[width=\linewidth]{figs/sup_mat/qualitative_scene/TRANSPORTER_v1_qualitative_scene-clay_grass.pdf}
        \caption{Gemma 3 modulations between \inlinebox{\texttt{clay}} and \inlinebox{\texttt{grass}}.}
        \label{fig:qualitative::scene:::clay} 
    \end{subfigure}
    \caption[]{\textbf{Examples of scene modulations}. Frame quality is compressed due to filesize (best viewed digitally).}
\end{figure*}
\begin{figure*}[ht]\ContinuedFloat
    \centering
    \begin{subfigure}[t]{\textwidth}
        \centering
        \includegraphics[width=\linewidth]{figs/sup_mat/qualitative_scene/TRANSPORTER_v1_qualitative_scene-tree_roof.pdf}
        \caption{Phi 4 MM modulations between \inlinebox{\texttt{tree}} and \inlinebox{\texttt{roof}}}.
        \label{fig:qualitative::scene:::tree}
    \end{subfigure}
    \begin{subfigure}[t]{\textwidth}
        \centering
        \includegraphics[width=\linewidth]{figs/sup_mat/qualitative_scene/TRANSPORTER_v1_qualitative_scene-overcast_clearsky.pdf}
        \caption{Phi 4 MM modulations between \inlinebox{\texttt{overcast}} and \inlinebox{\texttt{clear sky}}.}
        \label{fig:qualitative::scene:::sky}
    \end{subfigure}
    \caption[]{\textbf{Examples of scene modulations}. Frame quality is compressed due to filesize (best viewed digitally).}
\end{figure*}

\begin{figure*}[ht]
    \centering
    \label{fig:qualitative::multi}
    \begin{subfigure}[t]{\textwidth}
        \centering
        \includegraphics[width=\linewidth]{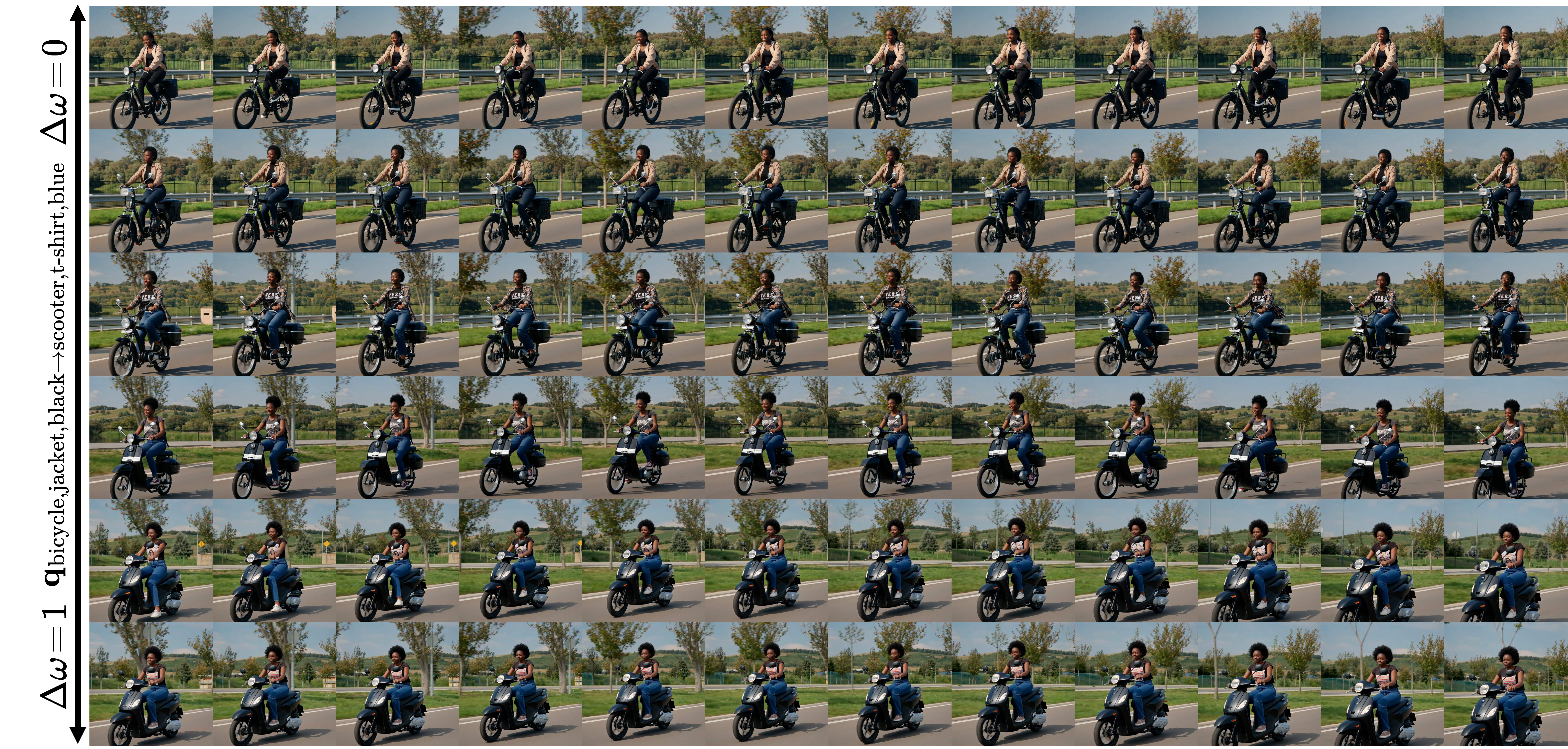}
        \caption{Video LLaMA 3 modulations between \inlinebox{\texttt{bicycle}},\inlinebox{\texttt{jacket}},\inlinebox{\texttt{black}} and \inlinebox{\texttt{scooter}},\inlinebox{\texttt{t-shirt}},\inlinebox{\texttt{blue}}.}
        \label{fig:qualitative::multi:::bike}
    \end{subfigure} 
    
    \begin{subfigure}[t]{\textwidth}
        \centering
        \includegraphics[width=\linewidth]{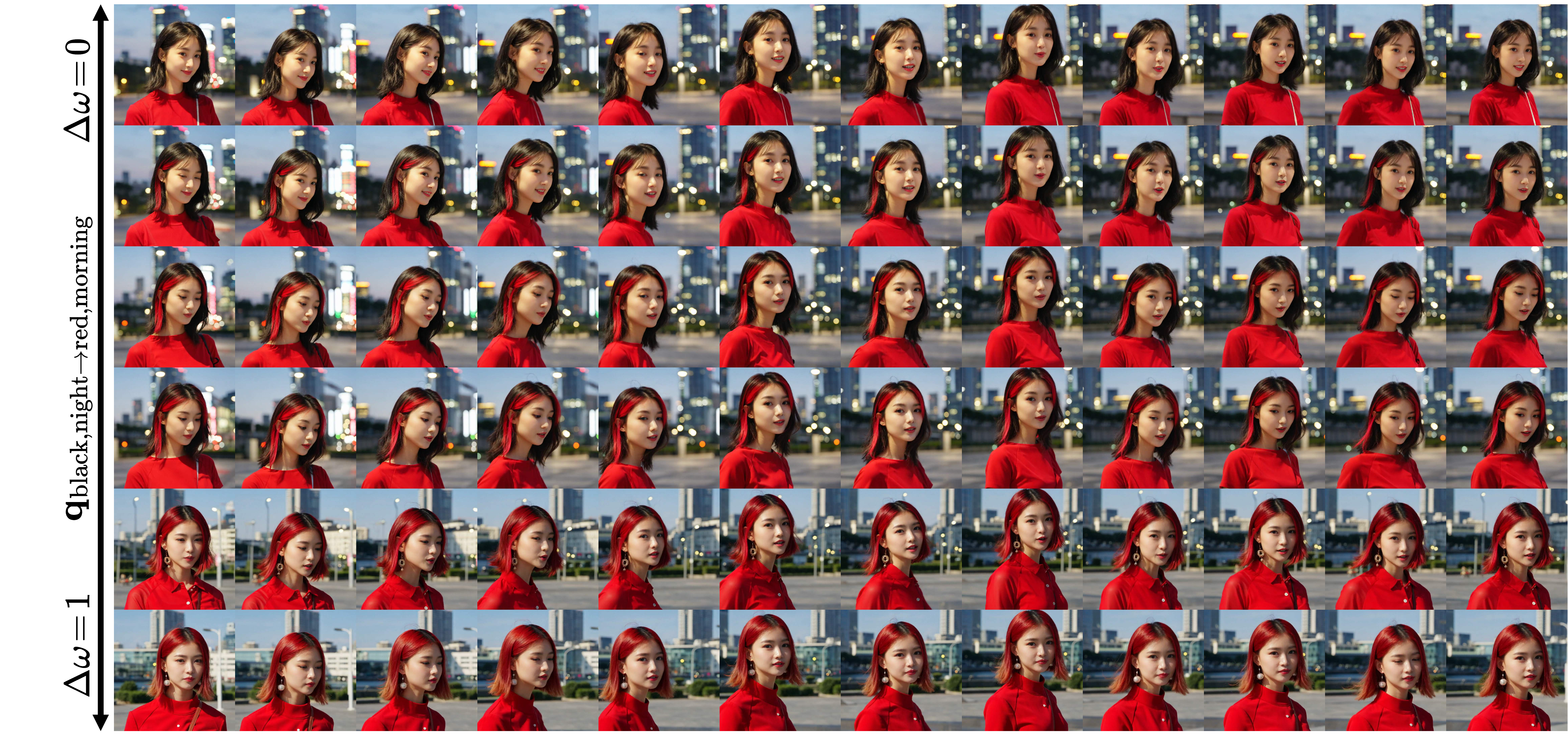}
        \caption{Gemma 3 modulations between \inlinebox{\texttt{black}},\inlinebox{\texttt{night}} and \inlinebox{\texttt{red}},\inlinebox{\texttt{morning}}.}
        \label{fig:qualitative::multi:::black}
    \end{subfigure}
    \caption{\textbf{Examples of multiple modulations}. Frame quality is compressed due to filesize (best viewed digitally).}
\end{figure*}
\begin{figure*}[ht]\ContinuedFloat
    \centering
    \begin{subfigure}[t]{\textwidth}
        \centering
        \includegraphics[width=\linewidth]{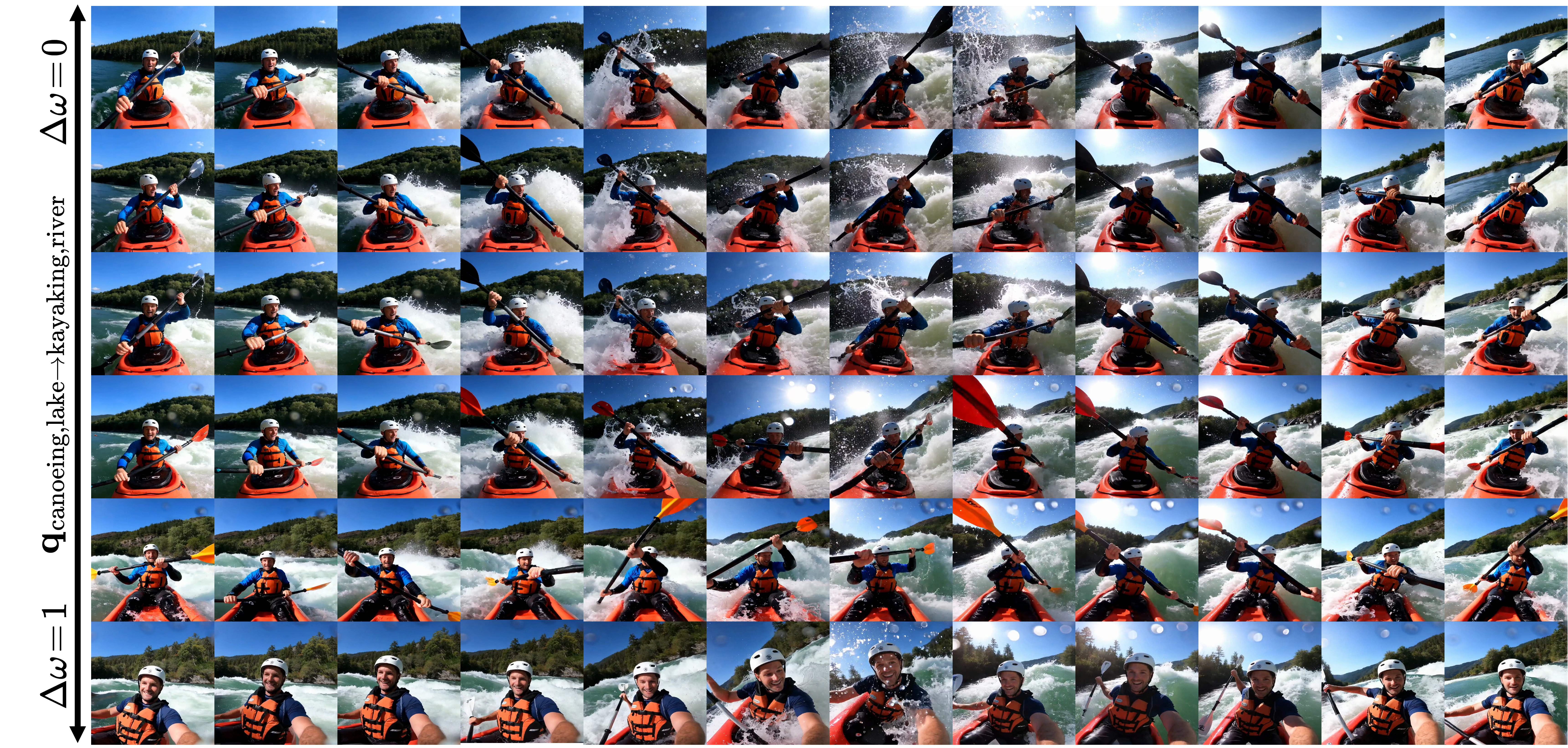}
        \caption{Phi 4 MM modulations between \inlinebox{\texttt{canoeing}},\inlinebox{\texttt{lake}} and \inlinebox{\texttt{kayaking}},\inlinebox{\texttt{river}}.}
        \label{fig:qualitative::multi:::canoe}
    \end{subfigure}
    \caption[]{\textbf{Examples of multiple modulations}. Frame quality is compressed due to filesize (best viewed digitally).}
\end{figure*}

\end{document}